\documentclass[runningheads]{llncs}
\usepackage[T1]{fontenc}
\usepackage{graphicx}
\usepackage{booktabs}
\usepackage[misc]{ifsym}
\newcommand{\corr}{(\Letter)}
\usepackage{mwe}

\usepackage{subfigure}
\usepackage{xcolor}
\usepackage{amsmath} 
\usepackage{amssymb}
\usepackage{multirow}
\usepackage{array}
\usepackage{bm}
\usepackage[colorlinks,bookmarksopen,bookmarksnumbered,citecolor=blue, linkcolor=blue, urlcolor=blue]{hyperref}

\definecolor{mycolor_1}{rgb}{0.0, 0.5, 0.9}
\definecolor{mycolor_2}{rgb}{0.9, 0.1, 0.5}
\definecolor{visualization_green}{rgb}{0.137, 0.545, 0.271}
\definecolor{visualization_red}{rgb}{0.851, 0.314, 0.298}
\newcommand{\ourmodel}{GLADMamba\xspace}

\begin{document}

\title{GLADMamba: Unsupervised Graph-Level Anomaly Detection Powered by Selective State Space Model}
\toctitle{GLADMamba: Unsupervised Graph-Level Anomaly Detection Powered by Selective State Space Model}
\tocauthor{Yali Fu, Jindong Li, Qi Wang, Qianli Xing}

\titlerunning{Published as a conference paper at ECML PKDD 2025}

\renewcommand{\thefootnote}{\fnsymbol{footnote}}
\author{Yali Fu\inst{1}\footnotemark[1]  \and
Jindong Li\inst{2}\footnotemark[1] \and
Qi Wang \inst{1,4} \corr \and
Qianli Xing \inst{3}
}

\renewcommand\thefootnote{}\footnotetext{%
\textsuperscript{$\star$} Equal contribution. 
}


\authorrunning{Published as a conference paper at ECML PKDD 2025}

\institute{School of Artificial Intelligence, Jilin University, Changchun, China \\  \email{fuyl23@mails.jlu.edu.cn, qiwang@jlu.edu.cn}
\and
Hong Kong University of Science and Technology (Guangzhou), Guangzhou, China \\  \email{jli839@connect.hkust-gz.edu.cn}
\and
College of Computer Science and Technology, Jilin University, Changchun, China   \\  \email{qianlixing@jlu.edu.cn}
\and 
Engineering Research Center of Knowledge-Driven Human-Machine Intelligence, Ministry of Education, China
}

\maketitle              

\begin{abstract}
Unsupervised graph-level anomaly detection (UGLAD) is a critical and challenging task across various domains, such as social network analysis, anti-cancer drug discovery, and toxic molecule identification. 
However, existing methods often struggle to capture long-range dependencies efficiently and neglect the spectral information. 
Recently, selective state space models, particularly Mamba, have demonstrated remarkable advantages in capturing long-range dependencies with linear complexity and a selection mechanism. 
Motivated by their success across various domains, we propose GLADMamba, a novel framework that adapts the selective state space model into UGLAD field. 
We design a View-Fused Mamba (VFM) module with a Mamba-Transformer-style architecture to efficiently fuse information from different graph views with a selective state mechanism. 
We also design a Spectrum-Guided Mamba (SGM) module with a Mamba-Transformer-style architecture to leverage the Rayleigh quotient to guide the embedding refinement process, considering the spectral information for UGLAD. 
GLADMamba can dynamically focus on anomaly-related information while discarding irrelevant information for anomaly detection. To the best of our knowledge, this is the first work to introduce Mamba and explicit spectral information to UGLAD. 
Extensive experiments on 12 real-world datasets demonstrate that GLADMamba outperforms existing state-of-the-art methods, achieving superior performance in UGLAD. The code is available at \href{https://github.com/Yali-Fu/GLADMamba}{https://github.com/Yali-Fu/GLADMamba}.

\end{abstract}

\section{Introduction}

Unsupervised graph-level anomaly detection (UGLAD) is a prevalent task in numerous real-world scenarios, including social network analysis, drug discovery, and toxic molecule identification~\cite{2023_ECMLPKDD_CVTGAD,2022_WSDM_GLocalKD,2021_TKDE_Survey_GLAD}. Its goal is to identify graphs that exhibit significantly different patterns from the majority, which often represent unexpected events or behaviors~\cite{2023_WSDM_GOOD-D,2022_Scientific-Reports_GLADC,2022_WSDM_GLocalKD}. Unlike supervised approaches, unsupervised methods don't require labeled data, making them more adaptable to real-world scenarios where labeled anomalies are scarce or costly to obtain. Despite the existence of many excellent methods, several challenges still exist in this field.

\begin{figure}[t!]
    \subfigure[Feature and structure.]{
        \includegraphics[width=0.4\textwidth]{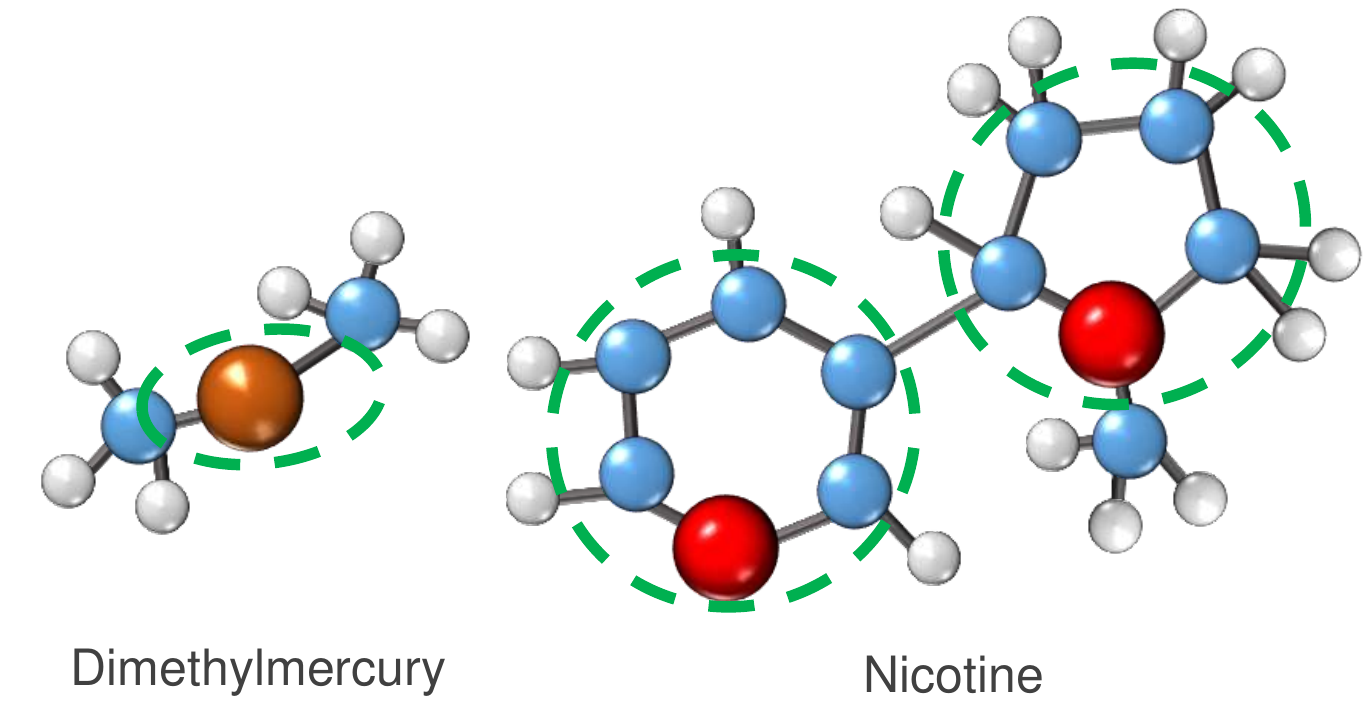}
        \label{fig:fig1_a}
    }
    \subfigure[The normalized spectral energy distributions.]{
        \includegraphics[width=0.27\textwidth]{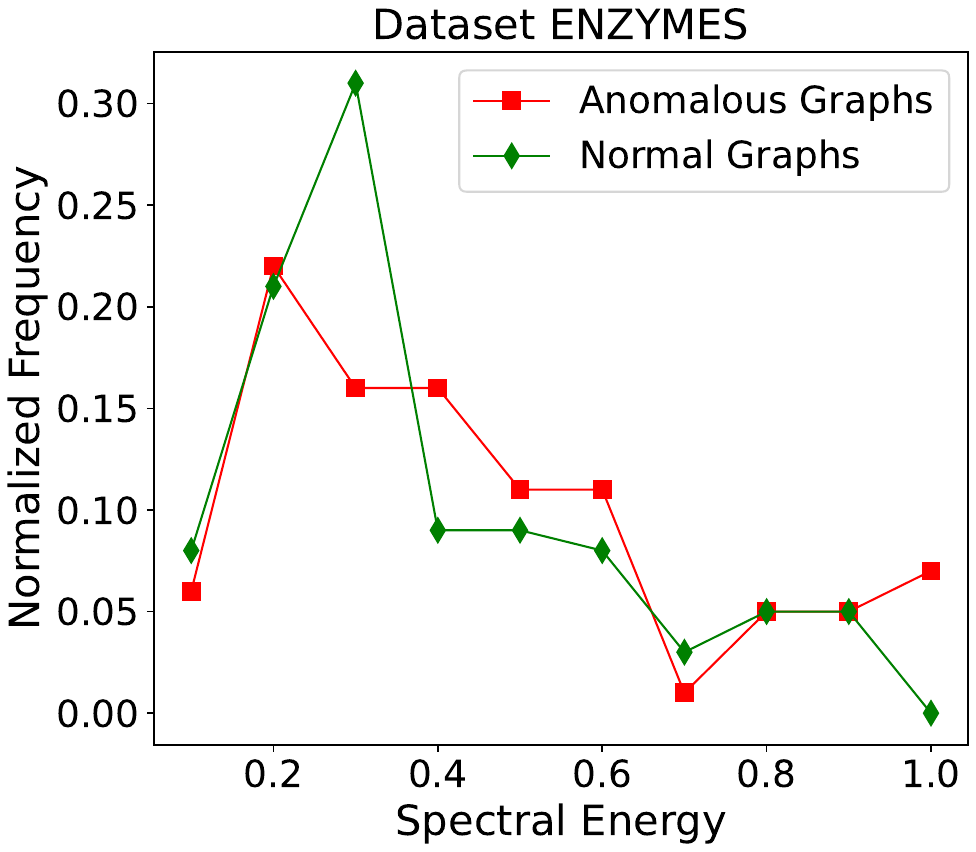}
        \includegraphics[width=0.27\textwidth]{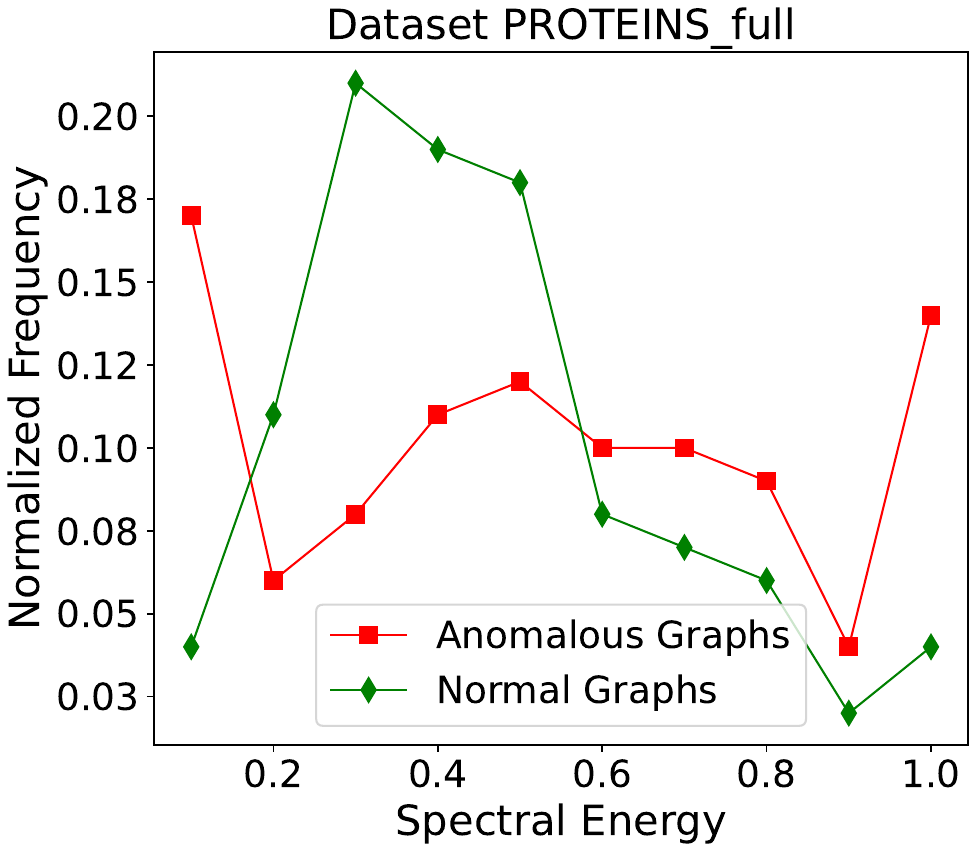}
        \label{fig:fig1_b}
    }
\caption{The key factors related to graph anomalies.}
\label{fig:fig_1}
\end{figure}

Most GNN-based methods are inherently limited by the over-squashing issue, which restricts their ability to effectively model long-range dependencies between nodes~\cite{2024_arXiv_Survey_Graph-Mamba,2024_arXiv(AAAI2025)_MOL-Mamba,2023_Survey_OverSmoothing_OverSquashing}. This limitation hampers the information propagation across distant nodes, making it challenging to capture anomaly-related patterns and ultimately weakening detection performance. Although some studies~\cite{2023_ECMLPKDD_CVTGAD,2025_AAAI_GeneralDyG,2024_CIKM_SI-HGAD} have incorporated Transformer into graph anomaly detection to mitigate this issue, the quadratic computational complexity of the attention mechanism significantly restricts the scalability of these methods, particularly for large-scale graphs. Additionally, as depicted in Fig.~\ref{fig:fig1_a}, relying solely on a single characteristic is insufficient for comprehensively capturing anomalies~\cite{2023_ECMLPKDD_CVTGAD,2023_WSDM_GOOD-D}. Thus, how to efficiently integrate information from multiple aspects (e.g., attributes and topologies) poses a critical challenge.

Furthermore, we observe spectral differences between normal and abnormal graphs in Fig.~\ref{fig:fig1_b}. And as demonstrated in ~\cite{2024_ICLR_RQGNN,2022_ICML_BWGNN,2024_AAAI_SEC-GFD,2025_AAAI_HIPGNN}, the energy distribution in the spectral domain shifts from low-frequency to high-frequency regions as the anomaly degree increases. However, most existing methods primarily focus on anomaly information in the spatial domain, neglecting the interaction with the spectral domain and failing to account for spectral differences between normal and anomalous graphs. Although a few works have explored the use of spectral information for graph-level anomaly detection, they rely on labeled data during training~\cite{2024_ICLR_RQGNN}. In the context of UGLAD, this remains an unexplored area, underscoring the need for further research and innovation.

Recently, selective state space models, especially Mamba~\cite{2024_COLM_Mamba}, originally designed for sequence modeling, have demonstrated remarkable advantages in capturing long-range dependencies with linear computational complexity and a selection mechanism. 
These advantages have been extensively validated across various domains~\cite{2024_arXiv_Survey_Graph-Mamba,2024_ISPL_Audio-Mamba,2024_arXiv_VL-Mamba,2024_arXiv_Survey_Vision-Mamba,2024_arXiv_Cobra}, making Mamba a strong candidate for addressing the challenges in UGLAD. Building on these strengths, we take the first step in integrating Mamba into UGLAD, unlocking its potential for more effective graph modeling while significantly improving computational efficiency and anomaly detection performance.

We propose a novel unsupervised graph-level anomaly detection framework that is powered by selective state space model (Mamba), named \ourmodel. First, we design a View-Fused Mamba (VFM) module, which is a Mamba-Transformer-style architecture and can efficiently fuse different graph views by Mamba's selective state transition mechanism. Benefiting from the architecture and advantages of Mamba, VFM excels at integrating multi-view information and capturing long-range dependencies, showcasing powerful embedding capabilities while maintaining linear time complexity. 
Furthermore, we design a Spectrum-Guided Mamba (SGM) module, which is also a Mamba-Transformer-style architecture and leverages explicit spectral information to guide the embedding refinement process for UGLAD. By the intrinsic relationship between spectral energy and graph anomalies, SGM establishes interactions between the spatial and spectral domains. Concretely, it employs the Rayleigh quotient to discretize the continuous state space of SGM, making system parameters spectrum-dependent. 
Therefore, the Rayleigh quotient can guide the update of latent states in SGM, enabling the model to selectively focus on anomaly-related information and filter redundant information to enhance detection performance.  

By the selection mechanism, \ourmodel can dynamically adjust its learning strategy based on specific characteristics of different input graphs, adaptively capturing anomaly patterns.
Extensive experiments on 12 real-world datasets demonstrate the effectiveness of \ourmodel. And \ourmodel consistently outperforms state-of-the-art methods in the unsupervised graph-level anomaly detection task.
Our key contributions are summarized as follows:
\begin{itemize}
    \item[$\bullet$] We propose a novel model, named \ourmodel, which adapts the selective state space model (Mamba) for unsupervised graph-level anomaly detection. To the best of our knowledge, this is the first work to introduce Mamba to UGLAD.
    \item[$\bullet$] We design a View-Fused Mamba (VFM) module for efficient multi-view fusion, boosting detection accuracy. In addition, we design a Spectrum-Guided Mamba (SGM) module, the first spectrum-guided method in this field to preserve anomaly-related patterns via selective state updates. This work pioneers spectrum-guided Mamba architecture for UGLAD.
    \item[$\bullet$] We conduct extensive experiments on 12 real-world datasets, demonstrating that \ourmodel achieves state-of-the-art performance in the unsupervised graph-level anomaly detection task.
\end{itemize}

\section{Related Work}

\subsection{Graph-Level Anomaly Detection}
Graph-level anomaly detection aims to identify anomalous graphs within a graph set, where anomalies typically represent rare but critical patterns compared to normal graphs~\cite{2024_arXiv_HC-GLAD,2021_TKDE_Survey_GLAD,2023_Big-Data_OCGIN}. Conventional methods generally involve two main steps: first, a graph kernel, such as the Weisfeiler-Lehman kernel (WL)~\cite{2011_JMLR_WL-graph-kernel} or propagation kernel (PK)~\cite{2016_ML_Propagation-kernels}, is used to learn representations; second, an anomaly detection algorithm, such as isolation forest (iF)~\cite{2008_ICDM_IF}, one-class support vector machine (OCSVM)~\cite{2001_JMLR_OCSVM}, or local outlier factor (LOF)~\cite{2000_SIGMOD_LOF}, is applied to detect anomalous graphs based on the extracted graph representations.

In addition, graph neural networks (GNNs)~\cite{2017_NeurIPS_Inductive-Representation-Learning-on-Large-Graphs,2016_arXiv_GCN,2017_arXiv_GAT,2018_arXiv_GIN} have attracted significant attention due to their remarkable performance in dealing with various graph data and tasks~\cite{2024_SSRN_CANNON,2022_Information-Sciences_CSPM,2022_arXiv_HGCL,2022_ICDM_CSPM}. Thus, various types of GNNs are employed as the backbone to conduct graph-level anomaly detection~\cite{2024_arXiv_HC-GLAD,2023_WSDM_GOOD-D,2022_Scientific-Reports_GLADC,2022_WSDM_GLocalKD,2023_Big-Data_OCGIN}. For example, 
GOOD-D~\cite{2023_WSDM_GOOD-D} designs a novel graph data augmentation method and employs contrastive learning at different levels for graph-level anomaly detection. 
CVTGAD~\cite{2023_ECMLPKDD_CVTGAD} employs a lightweight Transformer with an attention mechanism to model intra-graph and inter-graph node relationships, improving detection performance.

\subsection{State Space Models}

State Space Models (SSMs)~\cite{1960_SSM_first-paper} are classical frameworks for dynamic systems, while Structured SSMs (S4)~\cite{2022_ICLR_S4} enhance SSMs with efficient long-sequence modeling. Mamba~\cite{2024_COLM_Mamba,2024_arXiv_Survey_Mamba} builds on S4 by introducing a selection mechanism, enabling dynamic adaptation and improved efficiency. Together, they represent an evolution from traditional SSMs to modern, high-performance sequence modeling architectures. 
Beyond its core advancements, Mamba has demonstrated promising applications in various domains. It has been explored in computer vision~\cite{2024_arXiv_Survey_Vision-Mamba}, multimodal learning~\cite{2024_arXiv_VL-Mamba}, audio~\cite{2024_ISPL_Audio-Mamba}, and natural language processing~\cite{2024_arXiv_Cobra}, showcasing its versatility across different modalities.  
Mamba has also shown preliminary and promising applications in graph representation learning~\cite{2024_arXiv_Survey_Graph-Mamba}. For example, 
DG-Mamba~\cite{2024_arXiv(AAAI2025)_DG-Mamba} introduces a kernelized dynamic message-passing operator and a self-supervised regularization based on the principle of relevant information to improve efficiency, expressiveness, and robustness in dynamic graph learning.
MOL-Mamba~\cite{2024_arXiv(AAAI2025)_MOL-Mamba} enhances molecular representations by integrating hierarchical structural reasoning and electronic correlation learning, designing a hybrid Mamba-Graph and Mamba-Transformer framework supported by collaborative training strategies.

\section{Preliminaries}
\subsection{Problem Statement}

A graph is represented as $ G = (\mathcal{V}, \mathcal{E}, \mathbf{A}) $, where $\mathcal{V}$ denotes the set of nodes, $\mathcal{E}$ represents the set of edges, $\mathbf{A} \in \mathbb{R}^{|\mathcal{V}| \times |\mathcal{V}|}$ denotes the adjacency matrix, and $|\mathcal{V}|$ is the number of nodes.
The entry $\mathbf{A}_{i, j}$ of $\mathbf{A}$ is set to 1 if an edge exists between node $v_i$ and node $v_j$; otherwise, $\mathbf{A}_{i, j} = 0$. An attributed graph is defined as $G = (\mathcal{V}, \mathcal{E}, \mathbf{A}, \mathbf{F})$, where $\mathbf{F} \in \mathbb{R}^{|\mathcal{V}| \times d_f}$ is the feature matrix containing node attributes. Each row $\mathbf{f}$ of $\mathbf{F}$ corresponds to a feature vector of a node with $d_f$ dimensions. The collection of graphs is denoted as $\mathcal{G} = \{G_1, G_2, ..., G_m \}$, where $m$ is the total number of graphs.
This work addresses the GLAD problem under an unsupervised setting, where no labels are available for model training.

\subsection{State Space Models \& Mamba}
State Space Models (SSMs)~\cite{2024_COLM_Mamba,2022_ICLR_S4,2024_arXiv_Survey_Mamba} model the dynamic evolution of continuous systems via latent state \(h(t)\!\in\!\mathbb{R}^{N \times \tilde{L}}\), mapping input \(x(t)\!\in\!\mathbb{R}^{\tilde{L}}\) to output \(y(t)\!\in\!\mathbb{R}^{\tilde{L}}\) through the following state transition equation and observation equation:
\begin{equation}
    {h}^\prime(t) = \mathcal{A}{h}(t) + \mathcal{B}{x}(t),  \quad
    {y}(t) = \mathcal{C}{h}(t),
\label{Eq:ssm-1}
\end{equation}
where $\mathcal{A}\!\in\!\mathbb{R}^{N \times N}$ is the state transition matrix, $\mathcal{B}\!\in\!\mathbb{R}^{N \times 1}$ is the input matrix, $\mathcal{C}\!\in\!\mathbb{R}^{1 \times N}$ is the output matrix, $N$ denotes the state size, and $\tilde{L}$ denotes the sequence length.

By the step size $\Delta$, continuous parameters ($\mathcal{A}, \mathcal{B}$) are discretized into discrete forms ($\overline{\mathcal{A}}, \overline{\mathcal{B}}$) for practical application of SSMs. Discrete SSMs are described as follows:
\begin{equation}
    {h}_t = \overline{\mathcal{A}}{h}_{t-1} + \overline{\mathcal{B}}{x}_t, \quad
    {y}_t = \mathcal{C}{h}_t.
\label{Eq:ssm-2}
\end{equation}

The selective SSM (Mamba)~\cite{2024_COLM_Mamba} focuses on relevant information selectively by making parameters ($\mathcal{B}$, $\mathcal{C}$, $\Delta$) input-dependent.

\subsection{Rayleigh Quotient}
\label{Intro_Rayleigh_Quotient}

For a graph $G = (\mathcal{V}, \mathcal{E}, \mathbf{A})$, let $\mathbf{D}$ be the diagonal degree matrix with $\mathbf{D}_{ii} = \sum_j \mathbf{A}_{ij}$. Its Laplacian matrix $\bm{L}$ is defined as $\bm{L} = \mathbf{D} - \mathbf{A}$ (unnormalized) or $\bm{L} = \mathbf{I} - \mathbf{D}^{-1/2}\mathbf{A}\mathbf{D}^{-1/2}$ (normalized), where $\mathbf{I}$ is the identity matrix.  
The symmetric matrix $\bm{L}$ can be decomposed as $\bm{L} = \mathbf{U} \mathbf{\Lambda} \mathbf{U}^T$, where $\mathbf{\Lambda} = \text{diag}(\lambda_1, \lambda_2, \dots, \lambda_{|\mathcal{V}|})$ contains the corresponding eigenvalues sorted in ascending order, i.e., $0 \leq \lambda_1 \leq \lambda_2 \leq \dots \leq \lambda_{|\mathcal{V}|}$, and $\mathbf{U} = (\mathbf{u}_1, \mathbf{u}_2, \dots, \mathbf{u}_{|\mathcal{V}|})$ represents the orthonormal eigenvectors.  
Let $\bm{X} = (x_1, x_2, \dots, x_{|\mathcal{V}|})^{T}$ be the signal of $G$, and its graph Fourier transform is given by $\hat{\bm{X}} = (\hat{x}_1, \hat{x}_2, \dots, \hat{x}_{|\mathcal{V}|})^{T} = \mathbf{U}^T \bm{X}$.  
As demonstrated in~\cite{2024_ICLR_RQGNN,2022_ICML_BWGNN,2024_AAAI_SEC-GFD}, the Rayleigh quotient $R(\bm{L}, \bm{X})$ can represent the accumulated spectral energy, with a higher quotient value indicating more high-frequency components. We employ the following Rayleigh quotient without explicit eigenvalue decomposition for computational efficiency:
\begin{equation}
R(\bm{L}, \bm{X}) = \frac{\bm{X}^T \bm{L}\bm{X}}{\bm{X}^T \bm{X}} 
=\frac{\sum_{k=1}^{{|\mathcal{V}|}} \lambda_k \hat{x}_k^2}{\sum_{k=1}^{{|\mathcal{V}|}} \hat{x}_k^2}
= \frac{\sum_{(i,j)\in \mathcal{E}}(x_i - x_j)^2}{2\sum_{i\in \mathcal{V}} x_i^2}.
\label{Eq:RQ_Def}
\end{equation}

\begin{figure}[h!]
\centering
\subfigure[The overall pipeline of the proposed \ourmodel.]{
    \includegraphics[width=0.98\textwidth]{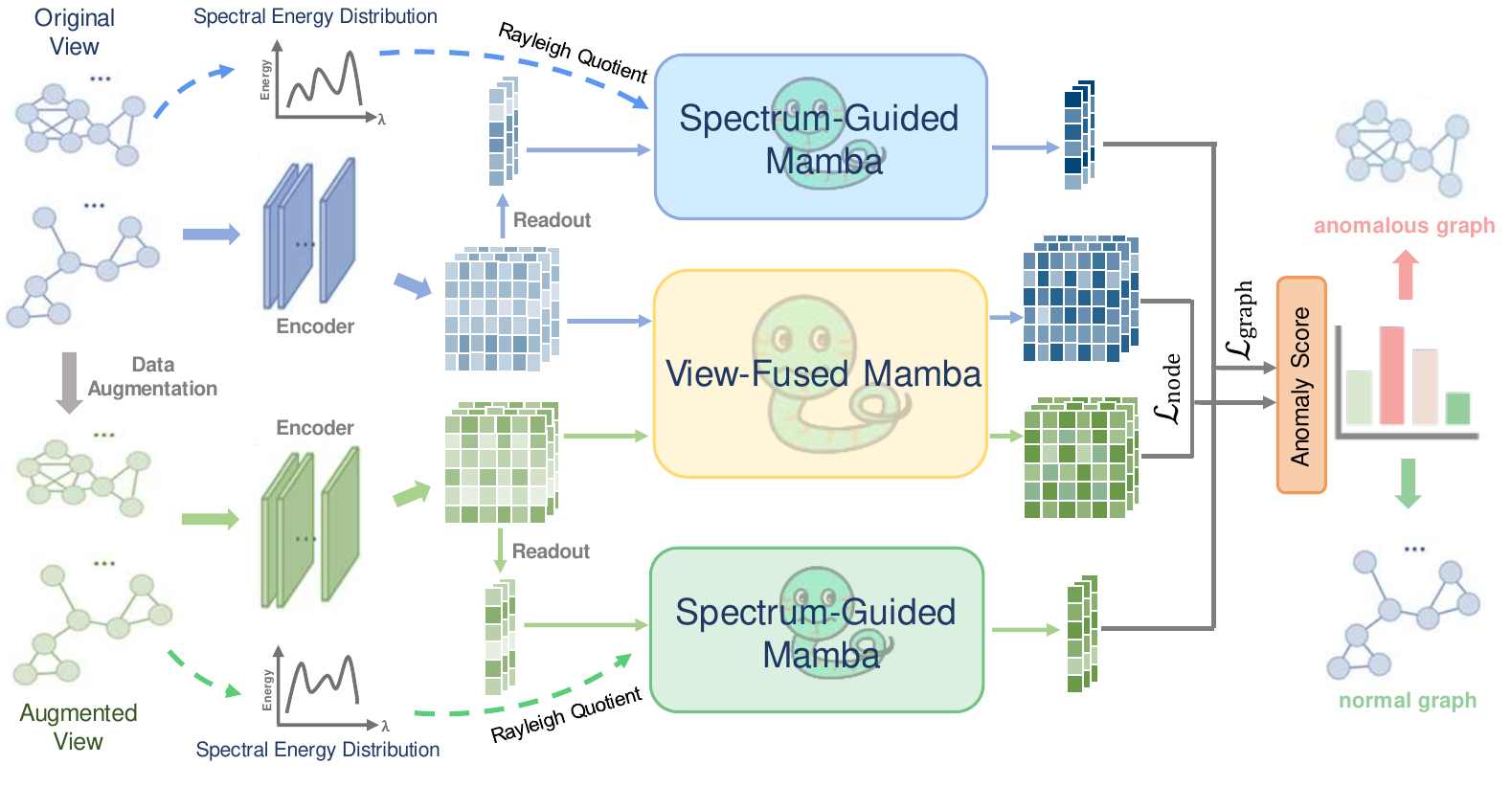}
    \label{fig:pipline}
}
\subfigure[View-Fused Mamba (VFM).]{
    \includegraphics[width=0.43\textwidth]{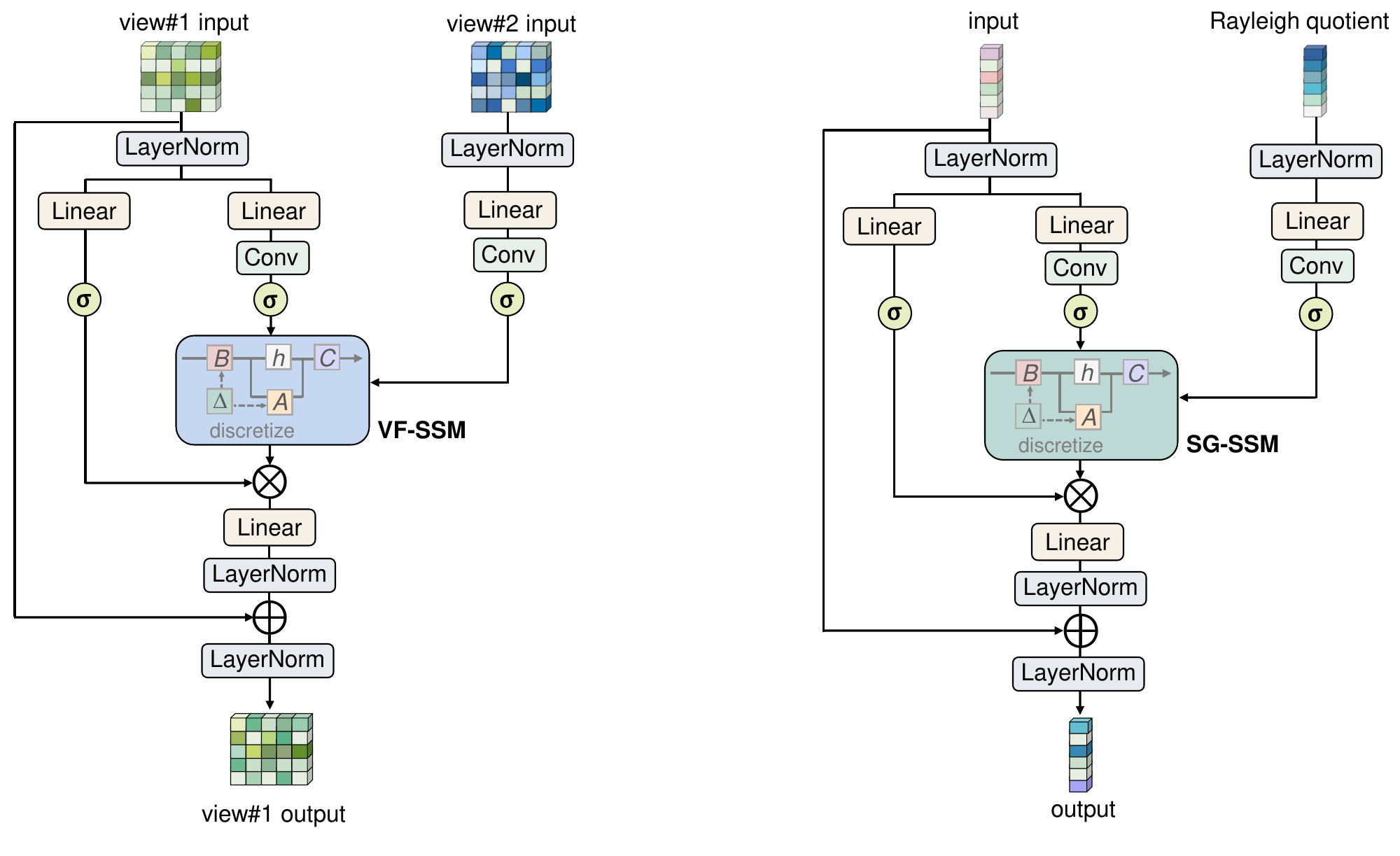}
    \label{fig:vf-mamba}
}
\hspace{0.5em}
\subfigure[Spectrum-Guided Mamba (SGM).]{
    \includegraphics[width=0.43\textwidth]{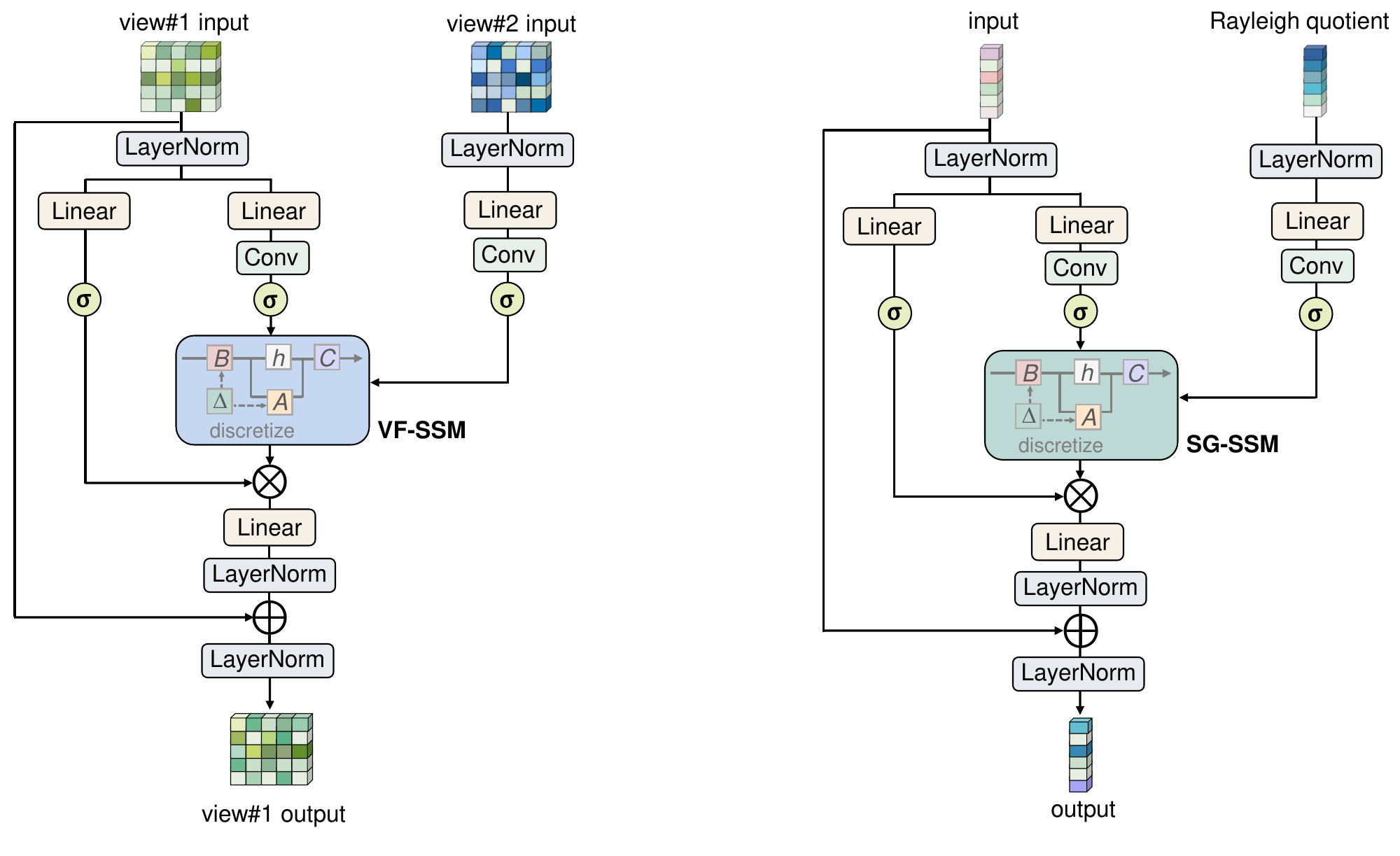}
    \label{fig:se-mamba}
}
\caption{Overview of the proposed \ourmodel model: (a) illustrates the overall pipeline of the framework; (b) depicts the View-Fused Mamba (VFM) module, which efficiently integrates multi-view information; (c) shows the Spectrum-Guided Mamba (SGM) module, which is designed to guide the embedding refinement process by Rayleigh quotient.}
\label{fig:framework}
\end{figure}

\section{Methodology}
\subsection{Data Augmentation and Encoding}
\label{Aug_Encoding}

For data augmentation, we obtain the original view $o$ and augmented view $a$ by a perturbation-free graph augmentation strategy~\cite{2023_ECMLPKDD_CVTGAD,2023_WSDM_GOOD-D,2023_AAAI_FedStar}, which is tailored for anomaly detection. And the views $o$ and $a$ focus on feature and structure characteristics of the graph, respectively.
After data augmentation, we employ two independent GNN encoders on two views to obtain node and graph embeddings. Taking the original view $o$ as an example, the GNN encoder updates node embeddings in the $l$-th layer according to the following message passing rule:

\begin{equation}
    \mathbf{h}_{v}^{(o, l)} = \text{UPDATE}^{(l-1)}\left(\mathbf{h}_{v}^{(o, l-1)}, \text{AGG}^{(l-1)}\left(\{\mathbf{h}_{u}^{(o,l-1)}: u \in \mathcal{N}(v)\}\right)\right),
\label{Eq:GNN}
\end{equation}
where $\mathbf{h}_v^{(o,l)}$ denotes the representation of node $v$ on the view $o$ at the $l$-th layer, $\mathcal{N}(v)$ represents the set of neighboring nodes of $v$, $\text{AGG}$ is the aggregation function to combine information from neighboring nodes, and $\text{UPDATE}$ is the update function to generate a new node representation. And we have $\mathbf{h}_v^{(o,0)}=\mathbf{f}_v$. After the $L$-layer GNN encoder, we obtain the representation $\mathbf{h}_{v}^{o}$ of node $v$ by concatenation operation and the representation $\mathbf{h}_{G}^{o}$ of graph $G$ on the view $o$ by the readout function:
\begin{equation}
    \mathbf{h}_{v}^{o} = [\mathbf{h}_{v}^{(o,1)}||\cdots||\mathbf{h}_{v}^{(o,L)}], \quad
    \mathbf{h}_{G}^{o} = \frac{1}{|\mathcal{V}_G|} \sum_{u \in \mathcal{V}_G} \mathbf{h}_u^{o},
\label{Eq:GNN_Emb}
\end{equation}
where $\mathcal{V}_G$ is the set of nodes in graph $G$. Let $\mathbf{H}^{o}$ denote the node representation matrix encoded from the original view $o$ in the training/testing batch, where each row corresponds to the representation of a node. In the same manner, we obtain the representation $\mathbf{h}_{v}^{a}$ of node $v$, the representation $\mathbf{h}_{G}^{a}$ of graph $G$, and the node representation matrix $\mathbf{H}^{a}$ for view $a$.

\subsection{View-Fused Mamba (VFM)}

In the View-Fused Mamba (VFM) module, we propose a novel view fusion mechanism. Specifically, we deeply fuse different aspects of graph data (i.e., feature and structure information) captured by the original view $o$ and augmented view $a$. Designed in a Mamba-Transformer-style architecture, the VFM efficiently captures and processes multi-view information for UGLAD. The detailed architecture of the View-Fused Mamba is depicted in Fig.~\ref{fig:vf-mamba}.

\textbf{Selective Parameterization in VF-SSM.}
For the parameter $\mathcal{A}$, we adopt HiPPO-LegS~\cite{2020_NeurlPS_HiPPO-LegS} for parameterization~\cite{2024_COLM_Mamba,2022_NeurlPS_S4D}. Before obtaining the input-dependent parameters ($\mathcal{B}$, $\mathcal{C}$, $\Delta$), we sequentially process the node representations $\mathbf{H}^{o}$ and $\mathbf{H}^{a}$ from two views through the following operations: a layer normalization~\cite{2016_LayerNorm}, a linear projection layer, a 1D convolutional layer, and a SiLU activation function. The formulas are as follows:

\begin{equation}
\mathbf{H}^{o}_{input} = \text{SiLU}(\text{Conv1D}(\text{Linear}(\text{LayerNorm}(\mathbf{H}^{o})))),
\label{Eq:node_norm_o}
\end{equation}
\begin{equation}
\mathbf{H}^{a}_{input} = \text{SiLU}(\text{Conv1D}(\text{Linear}(\text{LayerNorm}(\mathbf{H}^{a})))).
\label{Eq:node_norm_a}
\end{equation}

When parameterizing input-dependent parameters $\mathcal{B}$, $\mathcal{C}$ and $\Delta$, the two views are fused through the following formulas:
\begin{equation}
\mathcal{B}^{o} = \mathbf{W_{\mathcal{B}^{o}}}\mathbf{H}^{a}_{input},  \quad
\mathcal{C}^{o} = \mathbf{W_{\mathcal{C}^{o}}}\mathbf{H}^{a}_{input},  \quad
\Delta^{o} = \text{softplus}(\mathbf{W_{\Delta^{o}}}\mathbf{H}^{a}_{input}),
\label{Eq:BC_o}
\end{equation}
\begin{equation}
\mathcal{B}^{a} = \mathbf{W_{\mathcal{B}^{a}}}\mathbf{H}^{o}_{input}, \quad
\mathcal{C}^{a} = \mathbf{W_{\mathcal{C}^{a}}}\mathbf{H}^{o}_{input},  \quad
\Delta^{a} = \text{softplus}(\mathbf{W_{\Delta^{a}}}\mathbf{H}^{o}_{input}),
\label{Eq:BC_a}
\end{equation}
where $\mathbf{W}$ denotes the corresponding learnable matrix.

\textbf{Selective Discretization in VF-SSM.}
To discretize the parameters $\mathcal{A}$ and $\mathcal{B}$ into $\overline{\mathcal{A}}$ and $\overline{\mathcal{B}}$, we adopt the zero-order hold (ZOH) discretization rule, following~\cite{2024_COLM_Mamba,2024_arXiv_Dyg-mamba}.

\begin{equation}
    \begin{gathered}
        \overline{\mathcal{A}}^{o} = \exp(\Delta^{o} \mathcal{A}^{o}), \quad 
        \overline{\mathcal{B}}^{o} = (\Delta^{o} \mathcal{A}^{o})^{-1}(\exp(\Delta^{o} \mathcal{A}^{o}) - I)(\Delta^{o} \mathcal{B}^{o}),
    \end{gathered}
\label{Eq:ZOH_o}
\end{equation}
\begin{equation}
    \begin{gathered}
        \overline{\mathcal{A}}^{a} = \exp(\Delta^{a} \mathcal{A}^{a}), \quad
        \overline{\mathcal{B}}^{a} = (\Delta^{a} \mathcal{A}^{a})^{-1}(\exp(\Delta^{a} \mathcal{A}^{a}) - I)(\Delta^{a} \mathcal{B}^{a}).
    \end{gathered}
\label{Eq:ZOH_a}
\end{equation}

After the discretization of system parameters, the VF-SSM in VFM performs SSM($\overline{\mathcal{A}}^{o}$, $\overline{\mathcal{B}}^{o}$, $\mathbf{H}^{o}$) and SSM($\overline{\mathcal{A}}^{a}$, $\overline{\mathcal{B}}^{a}$, $\mathbf{H}^{a}$) to update state selectively following Eq.~(\ref{Eq:ssm-2}). The output of VF-SSM is as follows:
\begin{equation}
    \begin{gathered}
        \textbf{y}^{o}_{ssm} = \text{SSM}(\overline{\mathcal{A}}^{o}, \overline{\mathcal{B}}^{o},\mathbf{H}^{o}_{input}), \\
        \textbf{y}^{a}_{ssm} = \text{SSM}(\overline{\mathcal{A}}^{a}, \overline{\mathcal{B}}^{a},\mathbf{H}^{a}_{input}).
    \end{gathered}
\label{Eq:VF-SSM_output}
\end{equation}
After VFM, we obtain the final node representations $\mathbf{Z}^{o}$ and $\mathbf{Z}^{a}$ for the original and augmented views, respectively, as follows:
\begin{equation}
    \begin{gathered}
        \textbf{u}^{o} = \text{SiLU}(\text{Linear}(\text{LayerNorm}(\mathbf{H}^{o}))),\\
        \mathbf{Z}^{o} =  \text{LayerNorm}(\text{LayerNorm}(\text{Linear}(\textbf{y}^{o}_{ssm} \odot \textbf{u}^{o} )) + \mathbf{H}^{o}), \\
        \textbf{u}^{a} = \text{SiLU}(\text{Linear}(\text{LayerNorm}(\mathbf{H}^{a}))),\\
        \mathbf{Z}^{a} = \text{LayerNorm}(\text{LayerNorm}(\text{Linear}(\textbf{y}^{a}_{ssm} \odot \textbf{u}^{a} )) + \mathbf{H}^{a}),
    \end{gathered}
\label{Eq:VFMamba_output}
\end{equation}
where $\odot$ denotes element-wise multiplication.

\subsection{Spectrum-Guided Mamba (SGM)}

To consider spectral differences between normal and anomalous graphs for the UGLAD task, we design a specialized Spectrum-Guided Mamba (SGM) module, as illustrated in Fig.~\ref{fig:se-mamba}. The SGM module adopts the Rayleigh quotient as a measure of spectral characteristics, which is closely related to anomalies as described in Section~\ref{Intro_Rayleigh_Quotient}. Specifically, the SGM module utilizes the Rayleigh quotient to parameterize system parameters, making them spectrum-dependent and enabling spectrum-guided updates of latent states. Designed in a Mamba-Transformer-style architecture, the SGM effectively refines graph embeddings by selectively focusing on anomaly-relevant spectral information to enhance anomaly detection performance.

\textbf{Selective Parameterization in SG-SSM.}
Let $\mathbf{h}_{G}$ denote the graph representations encoded in Section~\ref{Aug_Encoding}. Firstly, we utilize the MLP to obtain the Rayleigh quotient representation $\mathbf{h}_{RQ}$, as follows:
\begin{equation}
\mathbf{h}_{RQ} = \text{MLP}(diag(R(\bm{L}, \bm{X})),
\label{Eq:RQ_Emb}
\end{equation}
where $diag(\cdot)$ extracts the diagonal elements of Rayleigh quotient $R(\bm{L}, \bm{X})$.

The inputs $\mathbf{h}_{G}$ and $\mathbf{h}_{RQ}$ are passed through a series of operations, including a layer normalization~\cite{2016_LayerNorm}, a linear projection, a 1D convolutional operation, and a SiLU activation function, as follows:

\begin{equation}
\mathbf{h}_{input} = \text{SiLU}(\text{Conv1D}(\text{Linear}(\text{LayerNorm}(\mathbf{h}_{G})))),
\label{Eq:graph_norm}
\end{equation}
\begin{equation}
\mathbf{h}_{RQ} = \text{SiLU}(\text{Conv1D}(\text{Linear}(\text{LayerNorm}(\mathbf{h}_{RQ})))).
\label{Eq:RQ_norm}
\end{equation}
We then utilize the obtained $\mathbf{h}_{RQ}$ to parameterize $\mathcal{B}$, $\mathcal{C}$ and $\Delta$, making system parameters $(\mathcal{B}, \mathcal{C}, \Delta)$ spectrum-dependent, as follows:
\begin{equation}
\mathcal{B} = \mathbf{W_{\mathcal{B}}}\mathbf{h}_{RQ}, \quad
\mathcal{C} = \mathbf{W_{\mathcal{C}}}\mathbf{h}_{RQ},  \quad
\Delta = \text{softplus}(\mathbf{W_{\Delta}}\mathbf{h}_{RQ}),
\label{Eq:BC_graph}
\end{equation}
where $\mathbf{W}$ is the corresponding learnable matrix.
For the parameter $\mathcal{A}$, we still adopt HiPPO-LegS~\cite{2020_NeurlPS_HiPPO-LegS} for parameterization~\cite{2024_COLM_Mamba,2022_NeurlPS_S4D}.

\textbf{Selective Discretization in SG-SSM.}
According to the ZOH rule, we discretize the parameters $\mathcal{A}$ and $\mathcal{B}$:
\begin{equation}
    \begin{gathered}
        \overline{\mathcal{A}} = \exp(\Delta \mathcal{A}), \quad
        \overline{\mathcal{B}} = (\Delta \mathcal{A})^{-1}(\exp(\Delta \mathcal{A}) - I)(\Delta \mathcal{B}).
    \end{gathered}
\label{Eq:ZOH_graph}
\end{equation}

The SG-SSM in SGM performs SSM($\overline{\mathcal{A}}$, $\overline{\mathcal{B}}$, $\mathbf{h}_{input}$) following Eq.~(\ref{Eq:ssm-2}), yielding the output $\textbf{y}_{ssm}$:
\begin{equation}
    \textbf{y}_{ssm} = \text{SSM}(\overline{\mathcal{A}}, \overline{\mathcal{B}},\mathbf{h}_{input}).
\label{Eq:SG-SSM_output}
\end{equation}
After the Spectrum-Guided Mamba, we obtain the final output $\mathbf{z}_{G}$ as follows:

\begin{equation}
    \begin{gathered}
        \textbf{u} = \text{SiLU}(\text{Linear}(\text{LayerNorm}(\mathbf{h}_{G}))),\\
        \mathbf{z}_{G} = \text{LayerNorm}(\text{LayerNorm}(\text{Linear}(y_{ssm} \odot \textbf{u})) + \mathbf{h}_{G}).
    \end{gathered}
\label{Eq:SGMamba_output}
\end{equation}

According to Eqs.~(\ref{Eq:RQ_Emb})-(\ref{Eq:SGMamba_output}), we obtain graph representations $\mathbf{z}^{o}_{G}$ and $\mathbf{z}^{a}_{G}$ for the original and augmented views, respectively.

\subsection{Training and Inference}
 
\textbf{Training.} 
We adopt the InfoNCE loss~\cite{2021_WWW_GCA} as contrastive objective to maximize the agreement between the representations from two views at node and graph scales: 

\begin{equation}
    \begin{gathered}
        \mathcal{L}^{'}_{node} = \frac{1}{|\mathit{B}|} \sum_{G_j \in \mathit{B}} \frac{1}{2|V_{G_j}|} \sum_{v_i \in V_{G_j}} \left[ \ell(\mathbf{z}_{i}^{o}, \mathbf{z}_i^{a}) + \ell(\mathbf{z}_i^{a}, \mathbf{z}_i^{o}) \right], \\
        \ell(\mathbf{z}_i^{o}, \mathbf{z}_i^{a}) = -\log \frac{e^{(cos(\mathbf{z}_i^{o}, \mathbf{z}_i^{a})/\tau)}}{\sum_{v_k \in V_{G_j} \setminus v_i} e^{(cos(\mathbf{z}_i^{o}, \mathbf{z}_k^{a})/\tau)}},
    \end{gathered}
\label{Eq:Loss_node_infoNce}
\end{equation}
where $\mathit{B}$ is the training batch, $\mathbf{z}_{i}^{o}$ and $\mathbf{z}_i^{a}$ are the embeddings of node $v_i$ on two views, $cos(,)$ is the cosine similarity function, and $\tau$ is the temperature parameter.
\begin{equation}
    \begin{gathered}
        \mathcal{L}_{graph}^{'} = \frac{1}{2|\mathit{B}|} \sum_{G_i \in \mathit{B}} \left[ \ell(\mathbf{z}_{G_i}^{o}, \mathbf{z}_{G_i}^{a}) + \ell(\mathbf{z}_{G_i}^{a}, \mathbf{z}_{G_i}^{o}) \right], \\ 
        \ell(\mathbf{z}_{G_i}^{o}, \mathbf{z}_{G_i}^{a}) = -\log \frac{e^{cos(\mathbf{z}_{G_i}^{o}, \mathbf{z}_{G_i}^{a})/\tau}}{\sum_{G_j \in \mathit{B} \setminus G_i} e^{cos(\mathbf{z}_{G_i}^{o}, \mathbf{z}_{G_j}^{a})/\tau}},
    \end{gathered}
\label{Eq:Loss_graph_infoNce}
\end{equation}
where $\mathbf{z}_{G_i}^{o}$ and $\mathbf{z}_{G_i}^{a}$ are the embeddings of graph $G_i$ on two views, and other notations are analogous to those in Eq.~(\ref{Eq:Loss_node_infoNce}).

During the training phase, we adopt an adaptive loss to consider different sensitivities of node and graph scales for different datasets~\cite{2023_ECMLPKDD_CVTGAD,2023_WSDM_GOOD-D}, as follows:
\begin{equation}
    \begin{gathered}
        \mathcal{L}_{node} = (\sigma_{node})^\alpha \mathcal{L}^{'}_{node},\quad
        \mathcal{L}_{graph} = (\sigma_{graph})^\alpha \mathcal{L}^{'}_{graph}, \\
        \mathcal{L} =  \mathcal{L}_{node} + \mathcal{L}_{graph},
    \end{gathered}
\label{Eq:Loss_adaptive}
\end{equation}
where $\alpha$ is the hyper-parameter and $\sigma$ is the standard deviation of predicted errors on the corresponding scale.

\textbf{Inference.} 
By minimizing $\mathcal{L}$ in Eq.~(\ref{Eq:Loss_adaptive}) during the training, the model learns common patterns of normal graphs. When testing an anomalous graph, the loss $\mathcal{L}$ tends to be significantly higher; therefore, we utilize $\mathcal{L}$ as the anomaly score. Additionally, the z-score standardization is adopted to balance anomaly scores from different scales. The final anomaly score is formulated as:
\begin{equation}
    {S} = \text{Std}(\mathcal{L}_{node}) + \text{Std}(\mathcal{L}_{graph}),
\label{Eq:Anomaly_Score}
\end{equation}
where Std($\mathcal{L}$) $=$ $(\mathcal{L} - \mu)$/${\sigma}$, and $\mu$ is the mean value of predicted errors of training samples at the corresponding scale.

\begin{table}[ht]
    \centering
    \caption{Statistics of the datasets~\cite{TuDataset} used in our experiments.}
    \label{tabs:datasets}
    \begin{tabular}{
        l |
        c  
        c 
        >{\centering\arraybackslash}p{2.1cm} 
        >{\centering\arraybackslash}p{1.9cm} 
        >{\centering\arraybackslash}p{2.0cm}
    }
        \toprule
        \textbf{Category} & \textbf{Dataset} & \textbf{Graphs} & \textbf{Avg. Nodes} & \textbf{Avg. Edges} & \textbf{Node Attr.} \\
        \midrule
        \multirow{1}{*}{Bioinformatics} 
            & ENZYMES        & 600   & 32.63   & 62.14   & 18 \\
        \midrule
        \multirow{9}{*}{Small molecules} 
            & AIDS           & 2000  & 15.69   & 16.20   & 4  \\
            & DHFR           & 467   & 42.43   & 44.54   & 3  \\
            & BZR            & 405   & 35.75   & 38.36   & 3  \\
            & COX2           & 467   & 41.22   & 43.45   & 3  \\
            & NCI1           & 4110  & 29.87   & 32.30   & -  \\
            & HSE            & 8417  & 16.89   & 17.23   & -  \\
            & MMP            & 7558  & 17.62   & 17.98   & -  \\
            & p53            & 8903  & 17.92   & 18.34   & -  \\
            & PPAR-gamma     & 8451  & 17.38   & 17.72   & -  \\
        \midrule
        \multirow{2}{*}{Social networks} 
            & IMDB-B         & 1000  & 19.77   & 96.53   & -  \\
            & REDDIT-B       & 2000  & 429.63  & 497.75  & -  \\
        \bottomrule
    \end{tabular}
\end{table}

\section{Experiments}

\subsection{Experiment Settings}
\textbf{Datasets.}
We conduct experiments on 12 public real-world datasets from TuDataset benchmark~\cite{TuDataset}, which involve small molecules, bioinformatics, and social networks. The statistics of the datasets are presented in Table~\ref{tabs:datasets}. Following the setting in~\cite{2023_ECMLPKDD_CVTGAD,2023_WSDM_GOOD-D,2022_WSDM_GLocalKD}, the samples in the minority class or real anomalous class are viewed as anomalies, while the rest are viewed as normal data. Only normal samples are used during training under the unsupervised setting.

\textbf{Baselines.}
To evaluate the effectiveness of \ourmodel, we compare it with 9 competitive baselines, spanning both earlier and recent approaches. These include the two-stage methods PK-iF~\cite{2008_ICDM_IF,2016_ML_Propagation-kernels}, WL-OCSVM~\cite{2001_JMLR_OCSVM,2011_JMLR_WL-graph-kernel}, WL-iF~\cite{2008_ICDM_IF,2011_JMLR_WL-graph-kernel}, InfoGraph-iF~\cite{2008_ICDM_IF,2020_ICLR_InfoGraph} and GraphCL-iF~\cite{2008_ICDM_IF,2020_NeurIPS_GraphCL}, as well as end-to-end methods OCGIN~\cite{2023_Big-Data_OCGIN}, GLocalKD~\cite{2022_WSDM_GLocalKD}, GOOD-D~\cite{2023_WSDM_GOOD-D} and CVTGAD~\cite{2023_ECMLPKDD_CVTGAD}. 
\begin{itemize}
    \item[$\bullet$] \textbf{Two-stage methods:} These methods first generate graph embeddings by graph kernels (e.g., propagation kernel (PK)~\cite{2016_ML_Propagation-kernels} or Weisfeiler-Lehman kernel (WL)~\cite{2011_JMLR_WL-graph-kernel}) or graph representation learning methods (e.g., InfoGraph \cite{2020_ICLR_InfoGraph} or GraphCL \cite{2020_NeurIPS_GraphCL}), and then apply traditional algorithms (e.g., isolation forest (iF)~\cite{2008_ICDM_IF} or one-class SVM (OCSVM)~\cite{2001_JMLR_OCSVM}) to identify anomalies.
    \item[$\bullet$] \textbf{End-to-end methods:} These approaches integrate graph representation learning and anomaly detection into a unified framework, enabling joint optimization. OCGIN~\cite{2023_Big-Data_OCGIN} adapts GIN for end-to-end graph anomaly detection by one-class classification objective. GLocalKD~\cite{2022_WSDM_GLocalKD} employs random distillation to learn normal patterns by training one GNN to predict another randomly-initialized GNN. GOOD-D~\cite{2023_WSDM_GOOD-D} uses hierarchical contrastive learning to detect anomalies through semantic inconsistency. CVTGAD~\cite{2023_ECMLPKDD_CVTGAD} employs a simplified Transformer for UGLAD.
\end{itemize}

\begin{table}[t!]
\caption{Overall performance comparison in terms of AUC (\%, mean±std). The best and second-best results are highlighted in \textcolor{mycolor_1}{\textbf{bold}} and \textcolor{mycolor_2}{\underline{underlined}}, respectively.}
\centering
\begin{tabular}{l|
                >{\centering\arraybackslash}p{1.8cm}
                >{\centering\arraybackslash}p{1.9cm}
                >{\centering\arraybackslash}p{1.8cm}
                >{\centering\arraybackslash}p{1.9cm}
                >{\centering\arraybackslash}p{1.9cm}}
\toprule[1.2pt]
Method         & PK-iF       & WL-OCSVM    & WL-iF       & InfoGraph-iF  & GraphCL-iF \\
\midrule
ENZYMES        & 51.30±2.01  & 55.24±2.66  & 51.60±3.81  & 53.80±4.50    & 53.60±4.88  \\
AIDS           & 51.84±2.87  & 50.12±3.43  & 61.13±0.71  & 70.19±5.03    & 79.72±3.98  \\
DHFR           & 52.11±3.96  & 50.24±3.13  & 50.29±2.77  & 52.68±3.21    & 51.10±2.35  \\
BZR            & 55.32±6.18  & 50.56±5.87  & 52.46±3.30  & 63.31±8.52    & 60.24±5.37  \\
COX2           & 50.05±2.06  & 49.86±7.43  & 50.27±0.34  & 53.36±8.86    & 52.01±3.17  \\
NCI1           & 50.58±1.38  & 50.63±1.22  & 50.74±1.70  & 50.10±0.87    & 49.88±0.53  \\
IMDB-B         & 50.80±3.17  & 54.08±5.19  & 50.20±0.40  & 56.50±3.58    & 56.50±4.90  \\
REDDIT-B       & 46.72±3.42  & 49.31±2.33  & 48.26±0.32  & 68.50±5.56    & 71.80±4.38  \\
HSE            & 56.87±10.51 & 62.72±10.13 & 53.02±5.12  & 53.56±3.98    & 51.18±2.71  \\
MMP            & 50.06±3.73  & 55.24±3.26  & 52.68±3.34  & 54.59±2.01    & 54.54±1.86  \\
p53            & 50.69±2.02  & 54.59±4.46  & 50.85±2.16  & 52.66±1.95    & 53.29±2.32  \\
PPAR-gamma     & 45.51±2.58  & 57.91±6.13  & 49.60±0.22  & 51.40±2.53    & 50.30±1.56  \\
\midrule
Avg.Rank       & 8.83        &7.50         & 8.58        & 6.83          & 7.42         \\
\bottomrule[1.2pt]
\end{tabular}

\begin{tabular}{l|cccc
                >{\centering\arraybackslash}p{2.2cm}}
\toprule[1.2pt]
Method         & OCGIN       & GLocalKD    & GOOD-D     & CVTGAD      & \ourmodel    \\
\midrule
ENZYMES        & 58.75±5.98  & 61.39±8.81  & 63.90±3.69  & \textcolor{mycolor_2}{\underline{67.79±5.43}} & \textcolor{mycolor_1}{\textbf{68.39±4.55}}  \\
AIDS           & 78.16±3.05  & 93.27±4.19  &97.28±0.69 & \textcolor{mycolor_1}{\textbf{99.39±0.55}} & \textcolor{mycolor_2}{\underline{99.29±0.47}}            \\
DHFR           & 49.23±3.05  & 56.71±3.57  & 62.67±3.11  & \textcolor{mycolor_2}{\underline{62.95±3.03}} & \textcolor{mycolor_1}{\textbf{63.79±4.16}} \\
BZR            & 65.91±1.47  & 69.42±7.78  & 75.16±5.15  & \textcolor{mycolor_2}{\underline{75.92±7.09}} & \textcolor{mycolor_1}{\textbf{77.25±4.62}}  \\
COX2           & 53.58±5.05  & 59.37±12.67 & 62.65±8.14  & \textcolor{mycolor_2}{\underline{64.11±3.22}} & \textcolor{mycolor_1}{\textbf{66.38±1.40}}  \\
NCI1           & \textcolor{mycolor_2}{\underline{71.98±1.21}} & 68.48±2.39  & 61.12±2.21  & 69.07±1.15  & \textcolor{mycolor_1}{\textbf{73.06±1.87}}  \\
IMDB-B         & 60.19±8.90  & 52.09±3.41  & 65.88±0.75  & \textcolor{mycolor_1}{\textbf{70.97±1.35}} & \textcolor{mycolor_2}{\underline{69.63±2.70}}           \\
REDDIT-B       & 75.93±8.65  & 77.85±2.62  & \textcolor{mycolor_1}{\textbf{88.67±1.24}} & 84.97±2.41  & \textcolor{mycolor_2}{\underline{86.12±0.41}}  \\
HSE            & 64.84±4.70  & 59.48±1.44  & 69.65±2.14  & \textcolor{mycolor_2}{\underline{70.30±2.90}} & \textcolor{mycolor_1}{\textbf{71.19±2.68}}  \\
MMP            & \textcolor{mycolor_2}{\underline{71.23±0.16}} & 67.84±0.59  & 70.57±1.56  & 70.96±1.01  & \textcolor{mycolor_1}{\textbf{73.19±3.22}}  \\
p53            & 58.50±0.37  & 64.20±0.81  & 62.99±1.55  & \textcolor{mycolor_2}{\underline{67.58±3.31}} & \textcolor{mycolor_1}{\textbf{68.38±1.62}}  \\
PPAR-gamma     & \textcolor{mycolor_1}{\textbf{71.19±4.28}} & 64.59±0.67  & 67.34±1.71  & 68.25±4.66  & \textcolor{mycolor_2}{\underline{69.21±3.46}}   \\
\midrule
Avg.Rank       & 4.50        & 4.58        & 3.25         & \textcolor{mycolor_2}{\underline{2.17}}              & \textcolor{mycolor_1}{\textbf{1.33}} \\
\bottomrule[1.2pt]
\end{tabular}
\label{tabs:Experiments_Overall-Performance}
\end{table}

\textbf{Evaluation Metrics.}
Following \cite{2023_ECMLPKDD_CVTGAD,2023_WSDM_GOOD-D,2022_WSDM_GLocalKD}, we use the area under the receiver operating characteristic curve (AUC) as the evaluation metric for UGLAD, where a higher AUC reflects better performance.

\textbf{Implementation Details.}
For the baselines, we report their public results from~\cite{2023_ECMLPKDD_CVTGAD,2023_WSDM_GOOD-D}.
We implement \ourmodel by PyTorch\renewcommand{\thefootnote}{1}\footnote{\noindent \href{https://pytorch.org/}{https://pytorch.org/}} on NVIDIA L40 and A40 GPUs, and ensure reproducibility by explicitly setting random seeds following \cite{2023_ECMLPKDD_CVTGAD,2023_WSDM_GOOD-D}. We employ GCN as the default GNN encoder. For AIDS, DHFR, HSE, and MMP datasets, the encoder employs GIN.

\subsection{Overall Performance Comparison}
We evaluate the performance of \ourmodel against several baselines in terms of AUC across 12 datasets. As summarized in Table~\ref{tabs:Experiments_Overall-Performance}, \ourmodel achieves the highest average rank, outperforming all baselines on 8 datasets and ranking second on the remaining datasets. The results also show that unified models surpass two-stage methods, highlighting the advantages of end-to-end optimization for representation learning and anomaly detection. These results underscore the effectiveness of \ourmodel in the UGLAD task across diverse domains.

\begin{figure}[t]
\centering
    \subfigure{
        \includegraphics[width=0.98\textwidth]{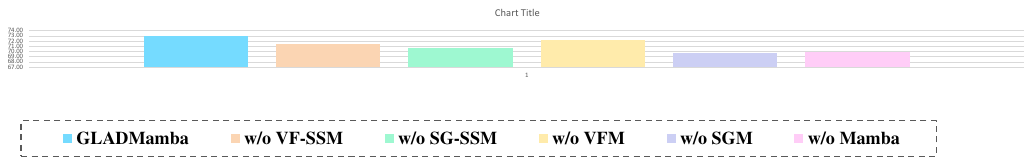}
    }
    \subfigure{
        \includegraphics[width=0.24\textwidth]{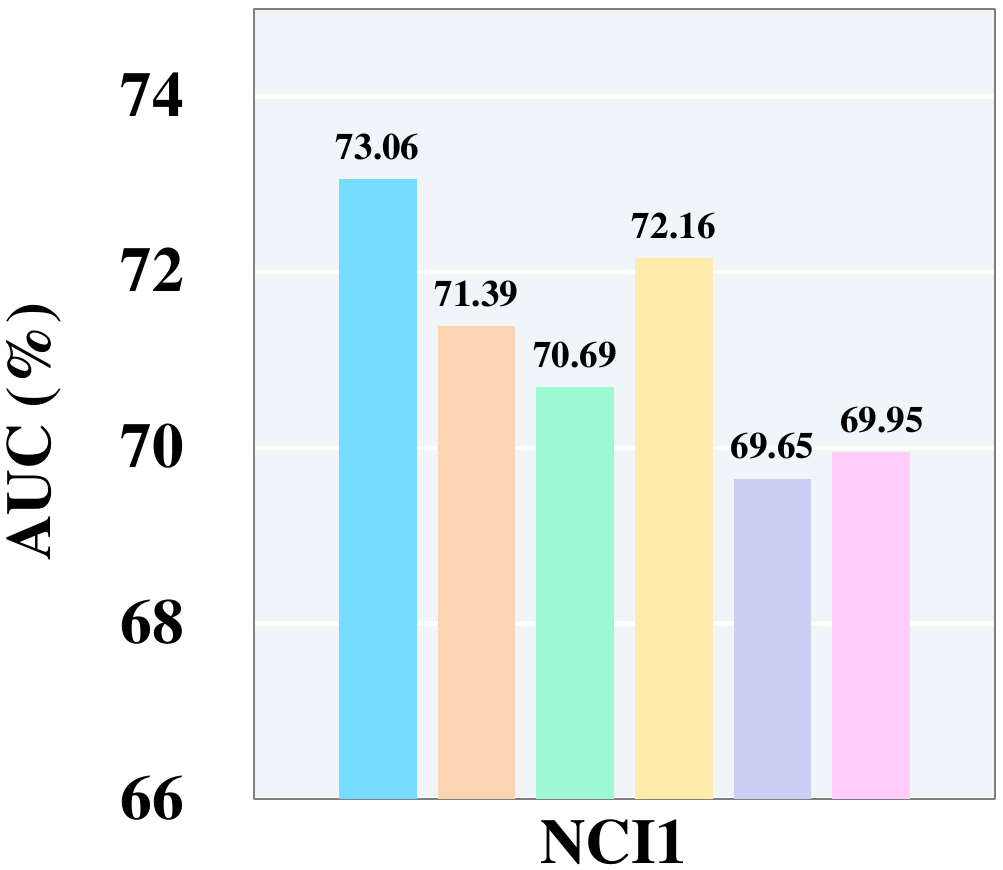}
    }
    \subfigure{
        \includegraphics[width=0.22\textwidth]{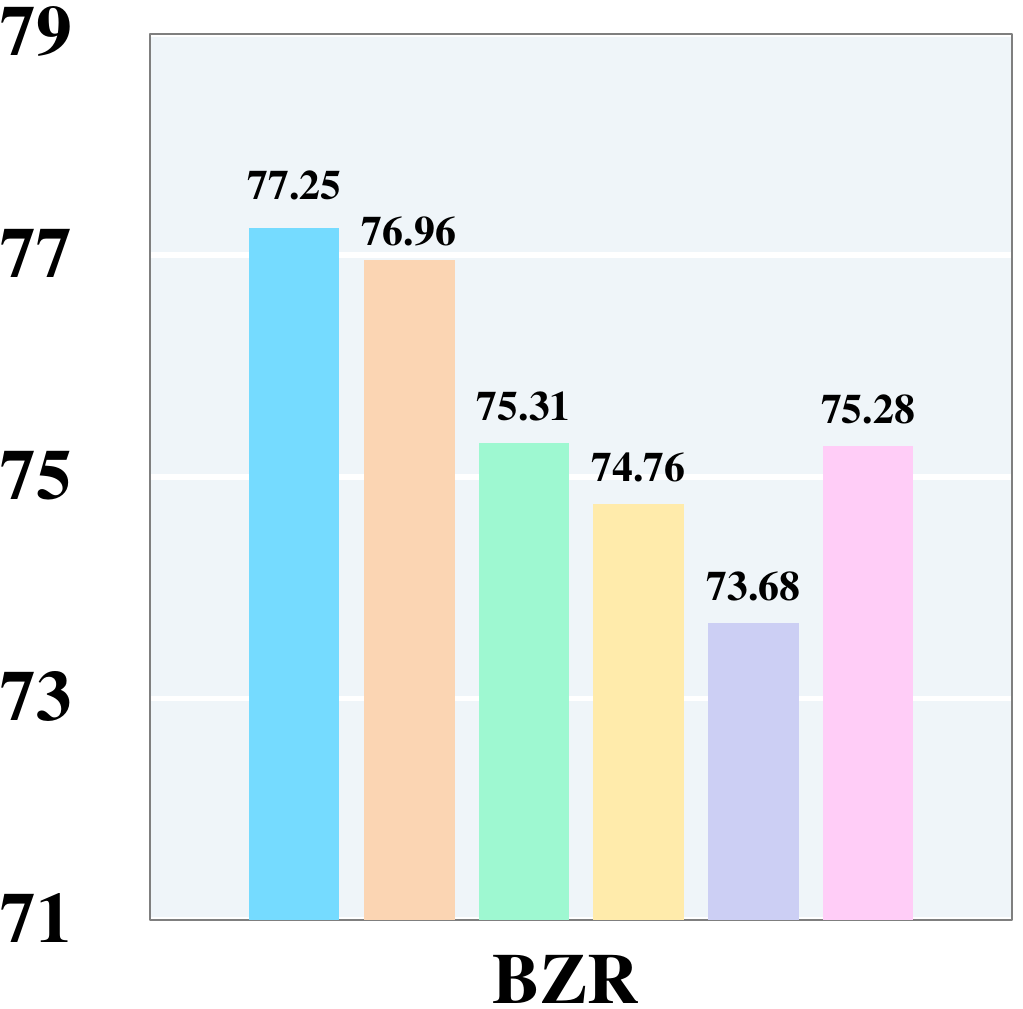}
    }
    \subfigure{
        \includegraphics[width=0.215\textwidth]{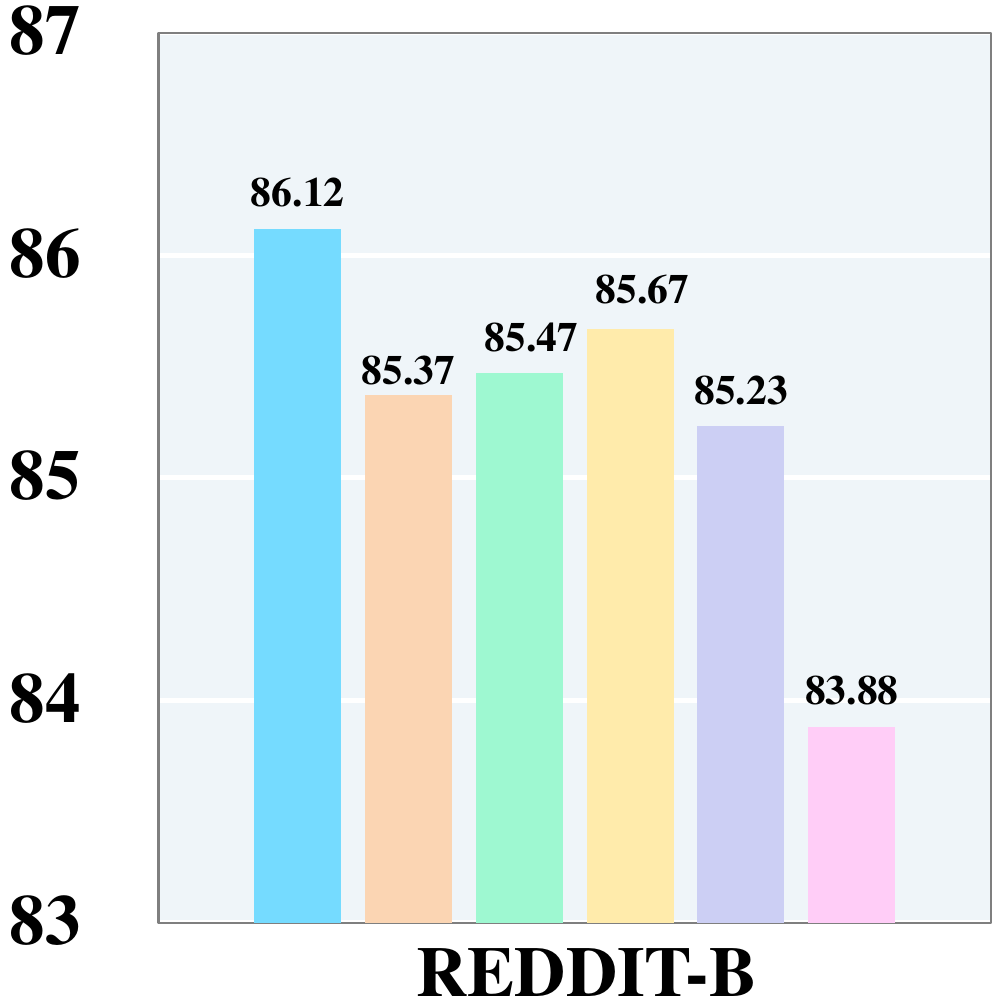}
    }
    \subfigure{
        \includegraphics[width=0.215\textwidth]{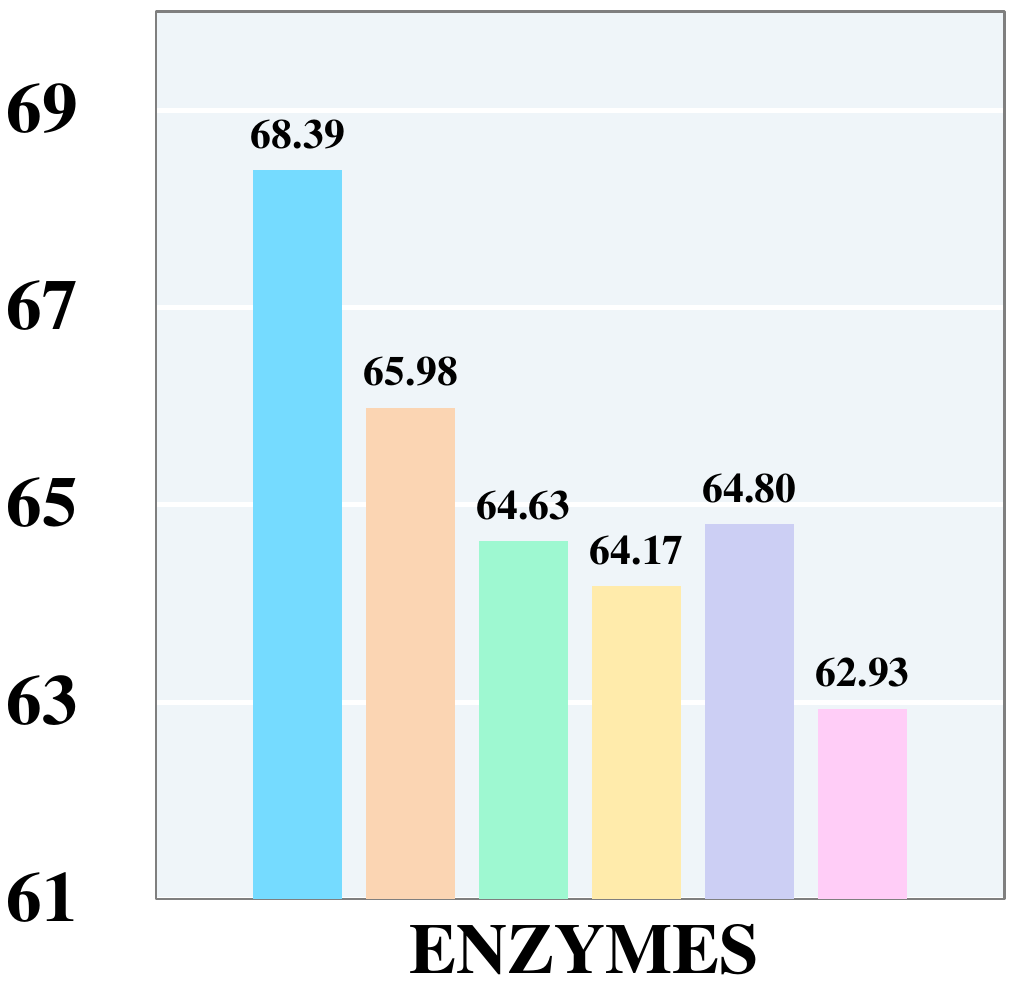}
    }
    \subfigure{
        \includegraphics[width=0.24\textwidth]{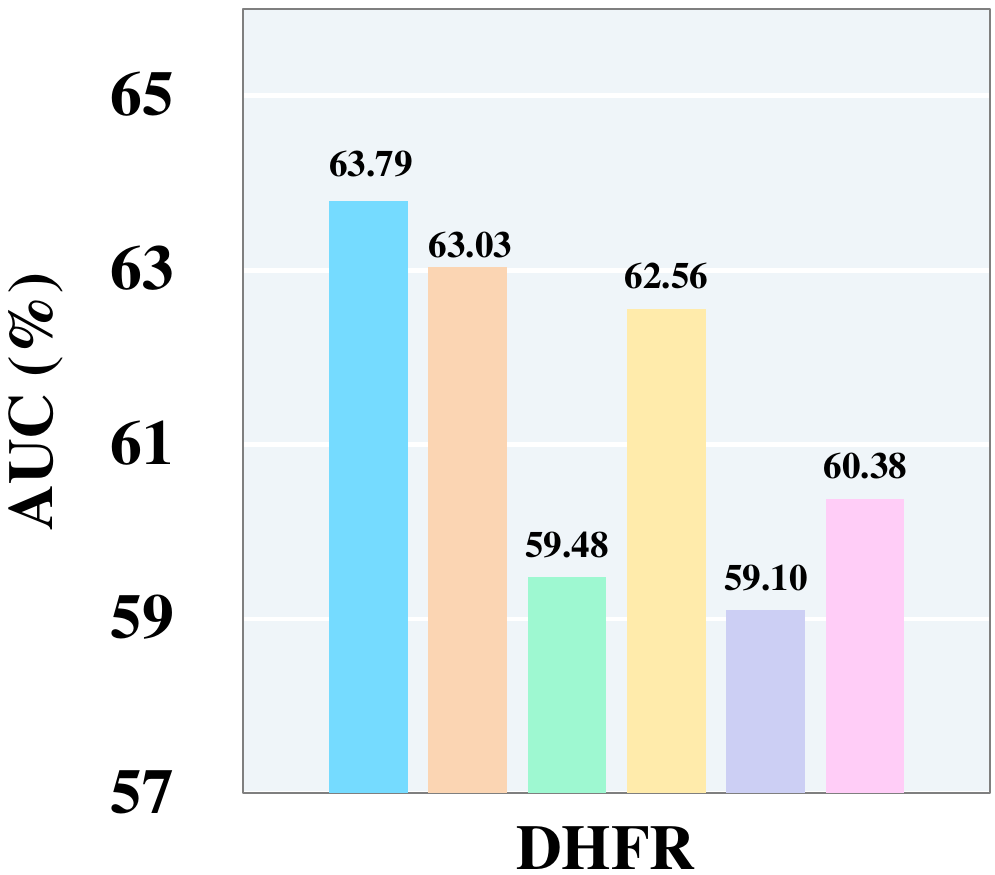}
    }
    \subfigure{
        \includegraphics[width=0.22\textwidth]{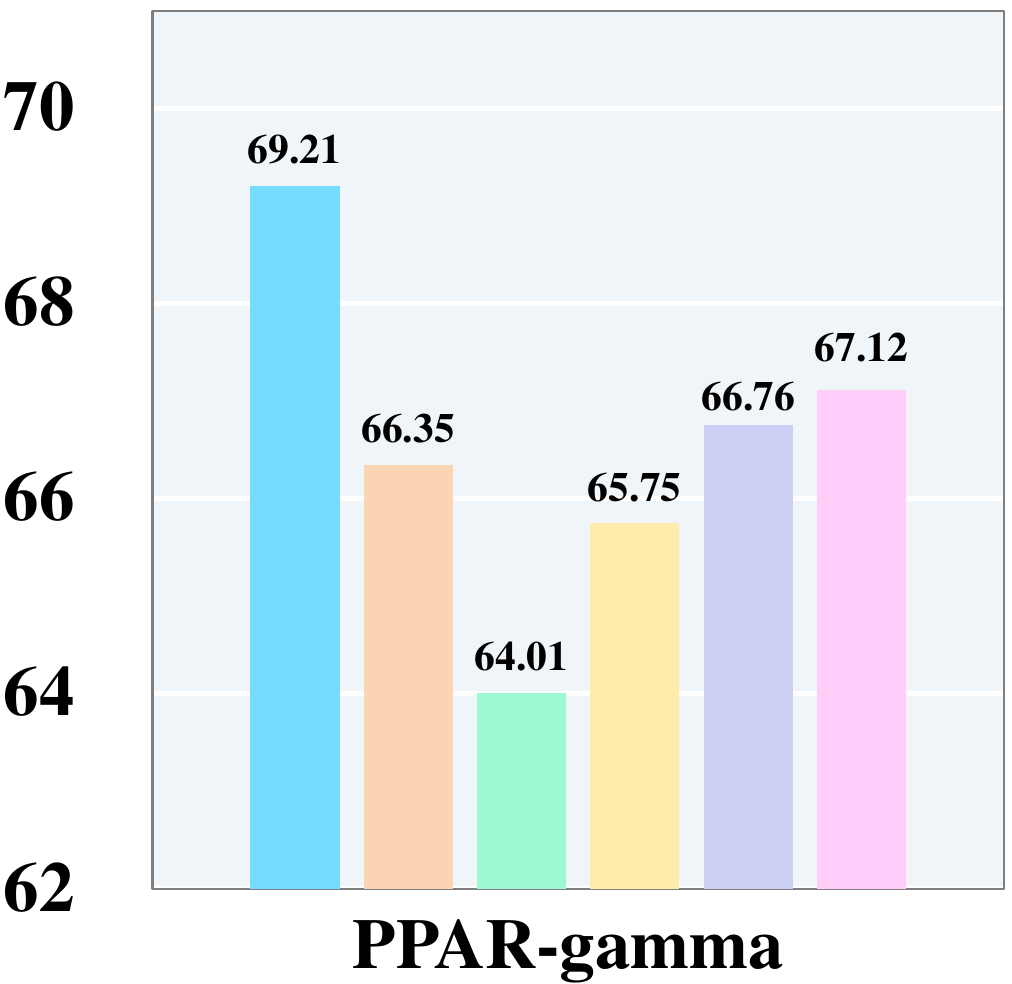}
    }
    \subfigure{
        \includegraphics[width=0.22\textwidth]{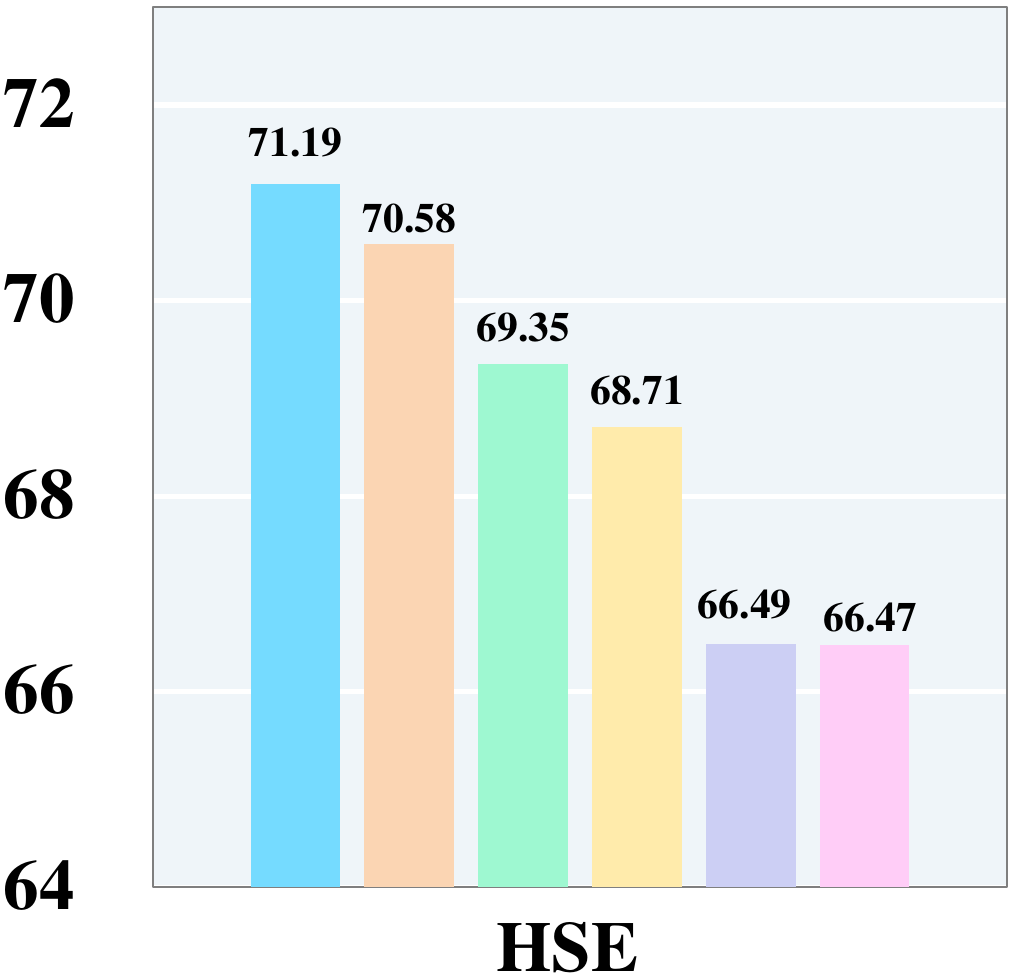}
    }
    \subfigure{
        \includegraphics[width=0.21\textwidth]{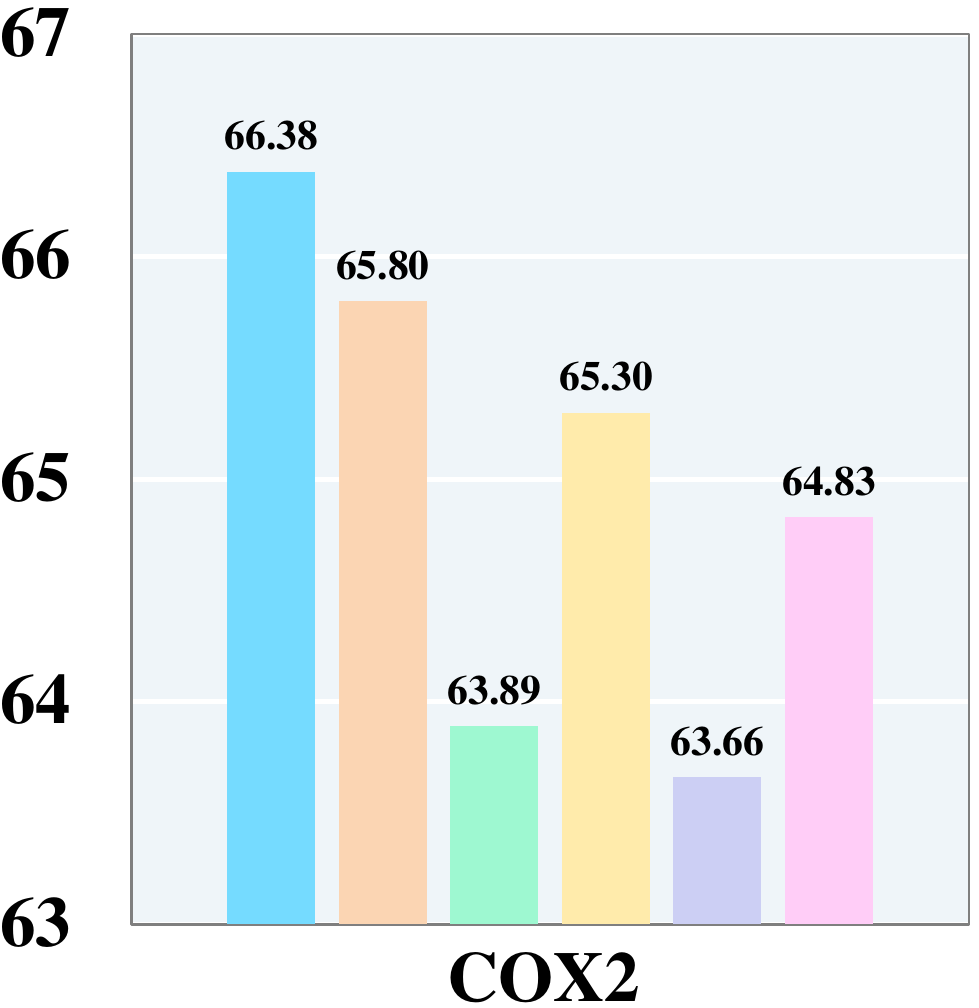}
    }

\caption{Ablation study on key components across representative datasets.} 
\label{Fig:Ablation}
\end{figure}

\subsection{Ablation Study}
To validate the effectiveness of key components of \ourmodel, we conduct extensive ablation experiments on 8 representative datasets in Fig.~\ref{Fig:Ablation}. The variant $w/o$ VF-SSM modifies the selective parameterization process in the VF-SSM using corresponding single-view inputs. The variant $w/o$ SG-SSM replaces the Rayleigh quotient with corresponding graph-level representations. The variants $w/o$ VFM, $w/o$ SGM, and $w/o$ Mamba denote the removal of the View-Fused Mamba, the Spectrum-Guided Mamba, and both, respectively. 

The results show that \ourmodel consistently outperforms all variants. The variants $w/o$ VFM, $w/o$ SGM, and $w/o$ Mamba significantly degrade performance, validating the effectiveness of Mamba in the UGLAD task and its superiority in capturing long-range dependencies. And the $w/o$ VF-SSM variant, which relies solely on single-view information, suffers a performance decline due to its limited ability to identify complex anomalies. Furthermore, the significant performance drop in the $w/o$ SG-SSM variant demonstrates the importance of spectral differences between normal and anomalous graphs for guiding \ourmodel's latent state updates. This also underscores that the SG-SSM component enhances \ourmodel's sensitivity to anomaly-related information by introducing Rayleigh quotient.

\begin{table}[b]
    \caption{Efficiency comparison on FLOPs, parameter size, and GPU memory usage.}
    \label{tabs:FLOPs-Params-GPU}
    \centering
    \begin{tabular}{l| >{\centering\arraybackslash}p{2.3cm} >{\centering\arraybackslash}p{2.3cm} >{\centering\arraybackslash}p{2.7cm} >{\centering\arraybackslash}p{2.3cm}}
        \toprule
        \textbf{Dataset} & \textbf{Model} & \textbf{FLOPs (M)$\downarrow$} &  \textbf{Params (MB)$\downarrow$} & \textbf{GPU (GB)$\downarrow$} \\
        \midrule
        \multirow{2}{*}{\centering \textbf{AIDS}} 
        & CVTGAD & 33.27 & 1.88 & 8.58 \\
        & GLADMamba & 48.97 & 0.07 & 5.12 \\
        \midrule
        \multirow{2}{*}{\centering \textbf{REDDIT-B}} 
        & CVTGAD & 602.46 & 13.19 & 30.70 \\
        & GLADMamba & 431.59 & 0.11 & 23.79 \\
        \midrule
        \multirow{2}{*}{\centering \textbf{p53}} 
        & CVTGAD & 701.31 & 20.61 & 4.98 \\
        & GLADMamba & 226.19 & 0.08 & 3.96 \\
        \bottomrule
    \end{tabular}
\end{table}

\subsection{Efficiency Analysis}
We assess the efficiency of \ourmodel on representative datasets in terms of FLOPs, parameter size, and GPU usage in Table~\ref{tabs:FLOPs-Params-GPU}. On the small-scale molecular dataset AIDS, the complexity differences between CVTGAD and \ourmodel are tolerable. However, on larger-scale datasets REDDIT-B and p53, the complexity differences become significant, particularly in terms of FLOPs and parameter size. This demonstrates the superiority of \ourmodel over Transformer-based approaches in scaling to large-scale graph anomaly detection.

\begin{figure}[t!]
\centering
\subfigure[The state size.]{
    \includegraphics[width=0.32\textwidth]{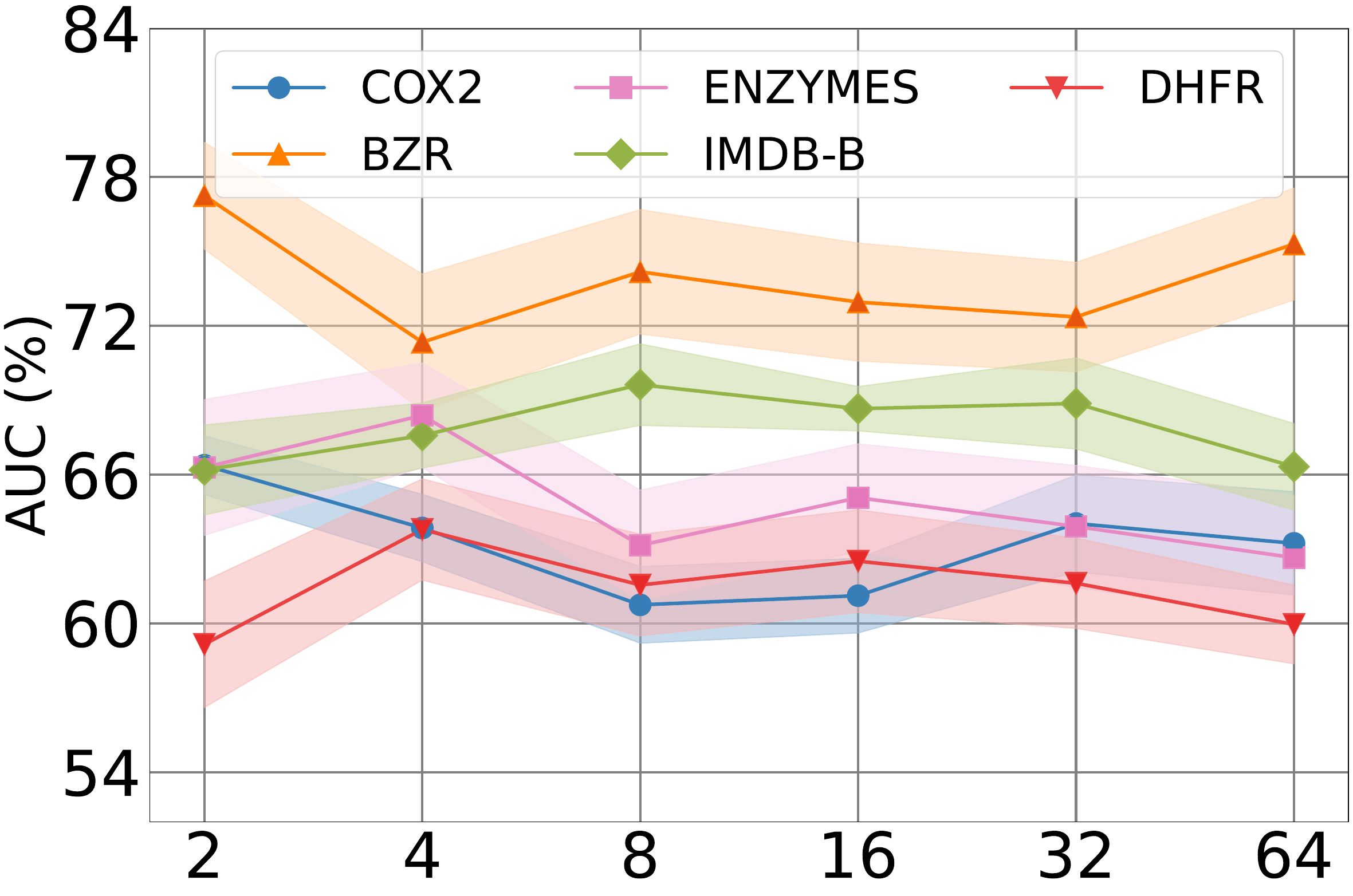}
    \label{Fig:Hyper-Para_state_size}
    }
\hspace{-1em}
\subfigure[The size of $\Delta$ projection.]{
    \includegraphics[width=0.32\textwidth]{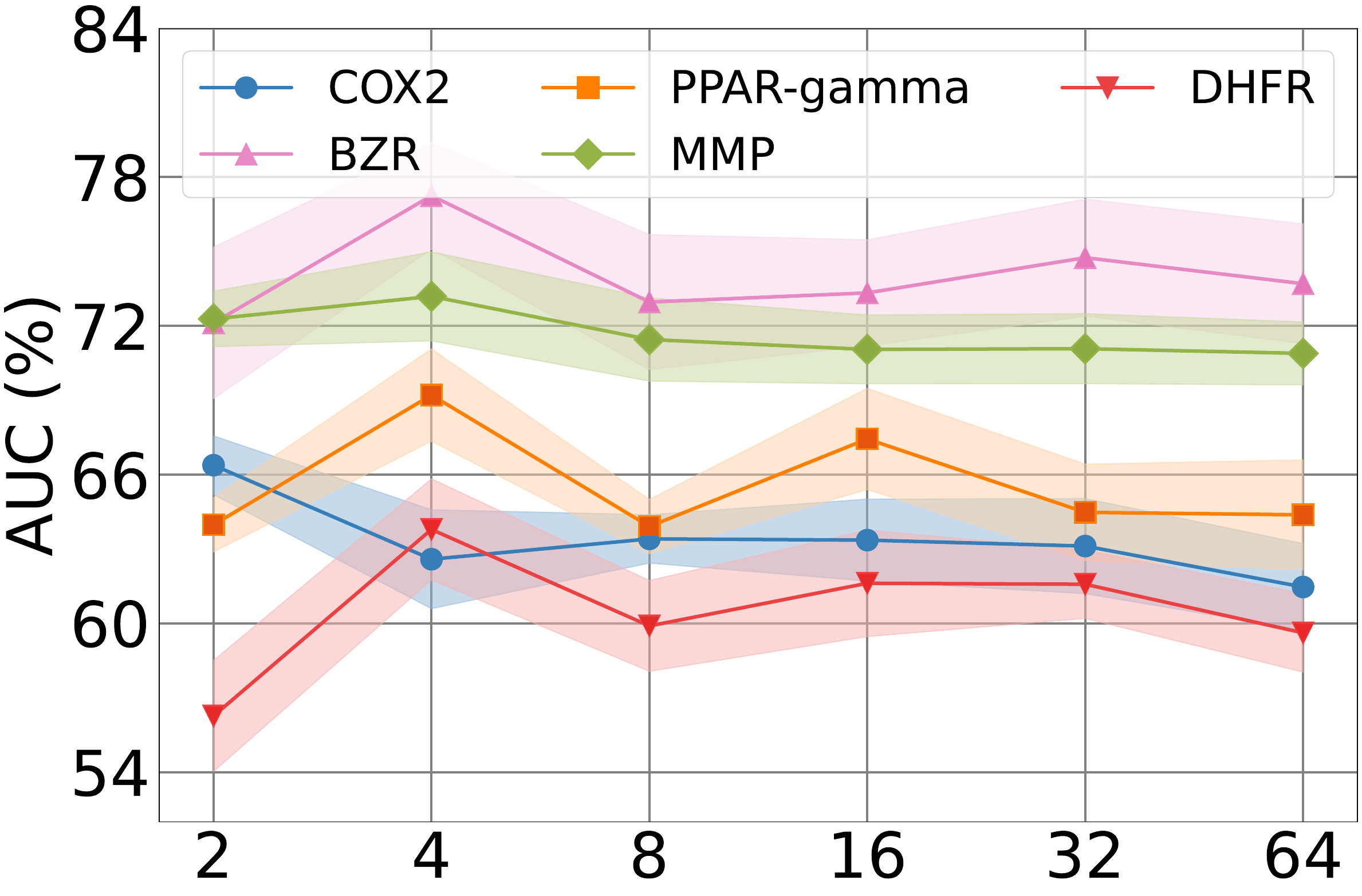}
    \label{Fig:Hyper-Para_rank}
    }
\hspace{-1em}
\subfigure[The convolution width.]{
    \includegraphics[width=0.32\textwidth]{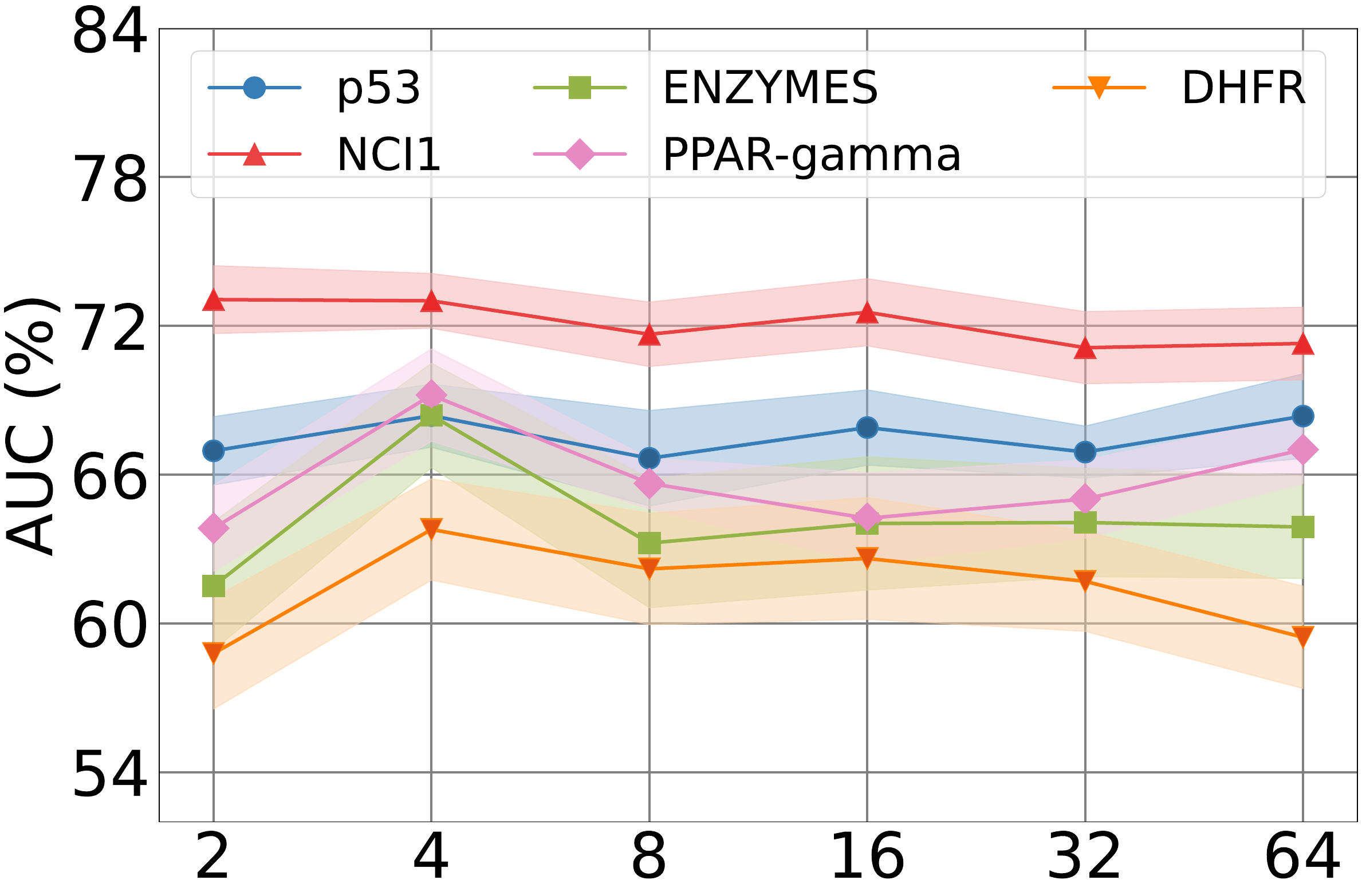}
    \label{Fig:Hyper-Para_conv_width}
    }
\caption{The hyper-parameter analysis on representative datasets.}
\label{fig:Hyper-Parameter}
\end{figure}

\subsection{Hyper-Parameter Analysis}
\textbf{The state size of VFM and SGM.}
To investigate the impact of state size in Mamba, we conduct hyper-parameter analysis experiments in Fig.~\ref{Fig:Hyper-Para_state_size}. The results indicate that performance variations with state size are not entirely consistent across datasets, likely due to the dependency of state size on dataset scale. Overall, the model can achieve satisfactory performance at smaller values (e.g., 4, 8).

\textbf{The size of $\Delta$ projection of VFM and SGM.}
The parameter $\Delta$ is constructed by linear projection of the input, controlling the strength of state updates. To investigate the impact of $\Delta$ projection size, we conduct experiments in Fig.~\ref{Fig:Hyper-Para_rank}. The results show that the model generally performs worst when the size is 2, except for the COX2 dataset. In general, the model can achieve respectable performance when the size is 4. As the projection size increases further, the performance remains relatively stable overall. These findings indicate that a moderate projection size (e.g., 4) is sufficient for optimal performance, avoiding suboptimal performance or unnecessary resource overhead.

\textbf{The local convolution width of VFM and SGM.}
The local convolution width refers to the kernel size of the Conv1D layer, controlling the model's receptive field. We conduct experiments to investigate its impact in Fig.~\ref{Fig:Hyper-Para_conv_width}.
We observe that when the width is 2, the performance degrades due to limited expressiveness of the Conv1D. In most cases, the model achieves optimal performance when the width is 4. Overall, GLADMamba remains relatively insensitive to this parameter, demonstrating robust performance.

\begin{figure}[t]
    \subfigure[Original view.]{
        \includegraphics[width=0.31\textwidth]{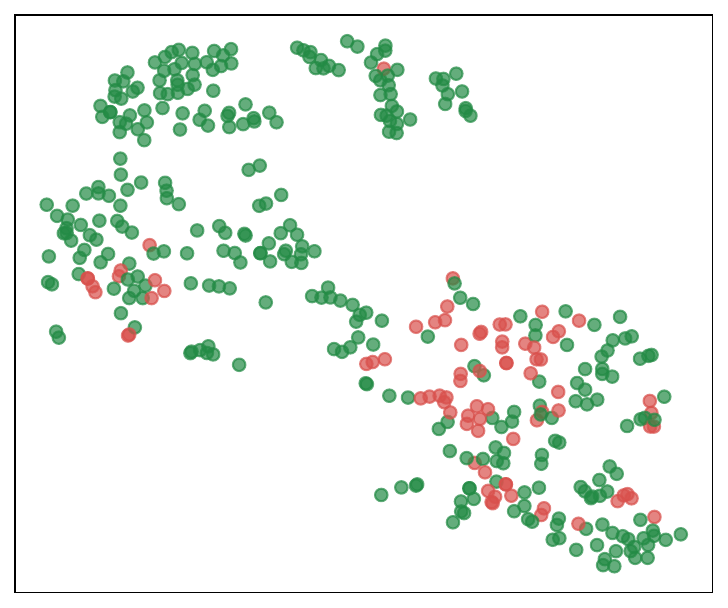}
    }
    \subfigure[Augmented view.]{
        \includegraphics[width=0.31\textwidth]{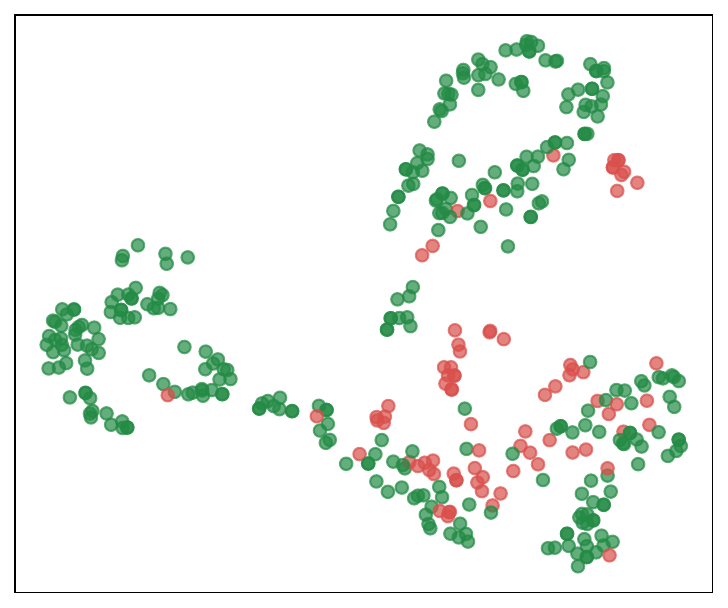}
    }
    \subfigure[Anomaly scores.]{
        \includegraphics[width=0.3\textwidth]{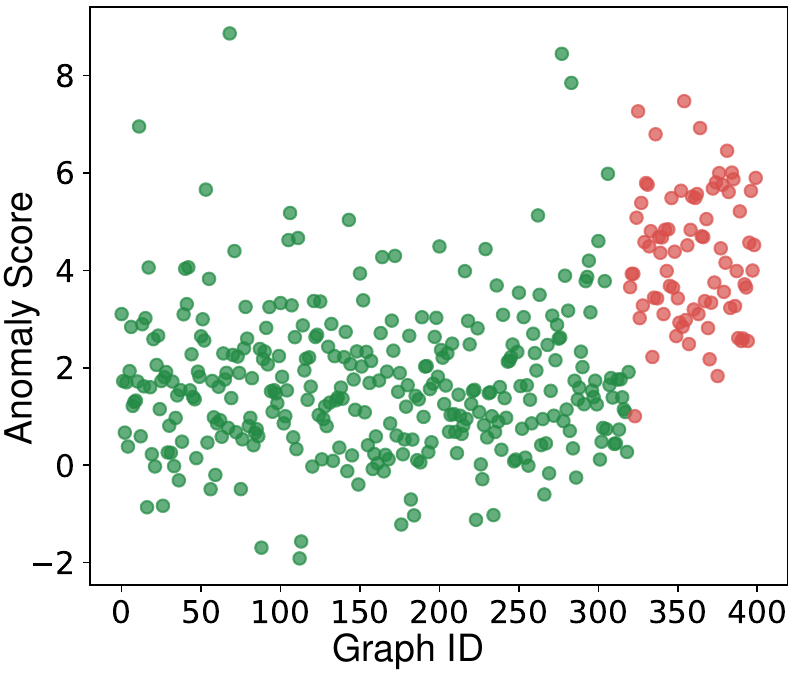}
    }
\caption{Visualization analysis on AIDS dataset. (\textcolor{visualization_green}{$\bullet$} normal graph, \textcolor{visualization_red}{$\bullet$} anomalous graph.)}
\label{fig:Visual}
\end{figure}

\subsection{Visualization Analysis}
We utilize t-SNE to visualize the graph embeddings from two views and anomaly scores learned by \ourmodel in Fig.~\ref{fig:Visual}. It can be observed that normal and anomalous graphs are not well-separated under a single view, as their distributions exhibit considerable overlap. However, when multiple views are integrated, normal and anomalous graphs exhibit a clear boundary in terms of anomaly scores, demonstrating the effectiveness of \ourmodel in UGLAD.

\section{Conclusion}

In this work, we introduce GLADMamba, a novel framework for UGLAD that effectively integrates the selective state space model and explicit spectral information. By leveraging the proposed View-Fused Mamba (VFM) and Spectrum-Guided Mamba (SGM) modules, \ourmodel dynamically selects and refines anomaly-related information, significantly improving detection performance. As far as we know, this is the first work to introduce Mamba and explicit spectral information to UGLAD. Experimental results on 12 real-world datasets demonstrate that GLADMamba outperforms existing state-of-the-art methods, highlighting its potential for advancing the UGLAD field. This fundamental architecture advancement establishes new possibilities for both graph learning and anomaly detection research.

\begin{credits}
\subsubsection{\ackname} This work is supported by the Youth Fund of the National Natural Science Foundation of China (No. 62206107).

\subsubsection{\discintname}
We declare that our research adheres to ethical guidelines, ensuring no harm or misuse of data and respecting privacy and confidentiality. We have conducted the study with integrity and disclosed any potential conflicts of interest.
\end{credits}

%
%
%
\bibliographystyle{splncs04}
%

\end{document}